\documentclass[10pt,twocolumn,letterpaper]{article}

\usepackage{cvpr}              

\usepackage[dvipsnames]{xcolor}

\usepackage{graphicx}
\usepackage{amsmath}
\usepackage{amssymb}
\usepackage{booktabs}
\usepackage{xcolor}
\usepackage{bbding}
\usepackage{mathrsfs}
\usepackage{wrapfig}
\usepackage{algorithm,algpseudocode}
\usepackage[accsupp]{axessibility} 

\newcommand{\norm}[1]{\left\lVert#1\right\rVert}
\newcommand{\e}[1]{\times 10^{#1}}

\newcommand{\dataset}{\textsc{DnD Group Gesture} }
\newcommand{\model}{\textsc{ConvoFusion} }

\definecolor{RDcolor}{rgb}{0.5, 0.1, 0.8}

\usepackage{soul}

\definecolor{bronze}{rgb}{1,1,0.6}
\definecolor{silve}{rgb}{0.969,0.796,0.600}
\definecolor{gold}{rgb}{0.941,0.592,0.600}
\newcommand{\gold}[1]{\colorbox{gold}{{#1}}}
\newcommand{\silver}[1]{\colorbox{silve}{{#1}}}

\usepackage[pagebackref,breaklinks,colorlinks]{hyperref}

\usepackage[capitalize]{cleveref}
\crefname{section}{Sec.}{Secs.}
\Crefname{section}{Section}{Sections}
\Crefname{table}{Table}{Tables}
\crefname{table}{Tab.}{Tabs.}

\def\paperID{4824} 
\def\confName{CVPR}
\def\confYear{2024}

\begin{document}

\title{ConvoFusion: Multi-Modal Conversational Diffusion\\for Co-Speech Gesture Synthesis
}
\author{Muhammad Hamza Mughal$^{1,2}$\hspace{1.8em} Rishabh Dabral$^{1}$\hspace{1.8em} 
Ikhsanul Habibie$^1$\hspace{1.8em} Lucia Donatelli$^3$\hspace{1.8em} \\
Marc Habermann$^{1}$\hspace{1.8em} Christian Theobalt$^{1,2}$\vspace{10pt}\\
$^1$Max Planck Institute for Informatics, SIC 
\hspace{1em}
$^2$Saarland University
\hspace{1em}
$^3$Vrije Universiteit Amsterdam\\
\\
}
\twocolumn[{ 
\renewcommand\twocolumn[1][]{#1} 
\maketitle 
\begin{center} 
    \vspace{-15pt} 
    \includegraphics[width=\textwidth]{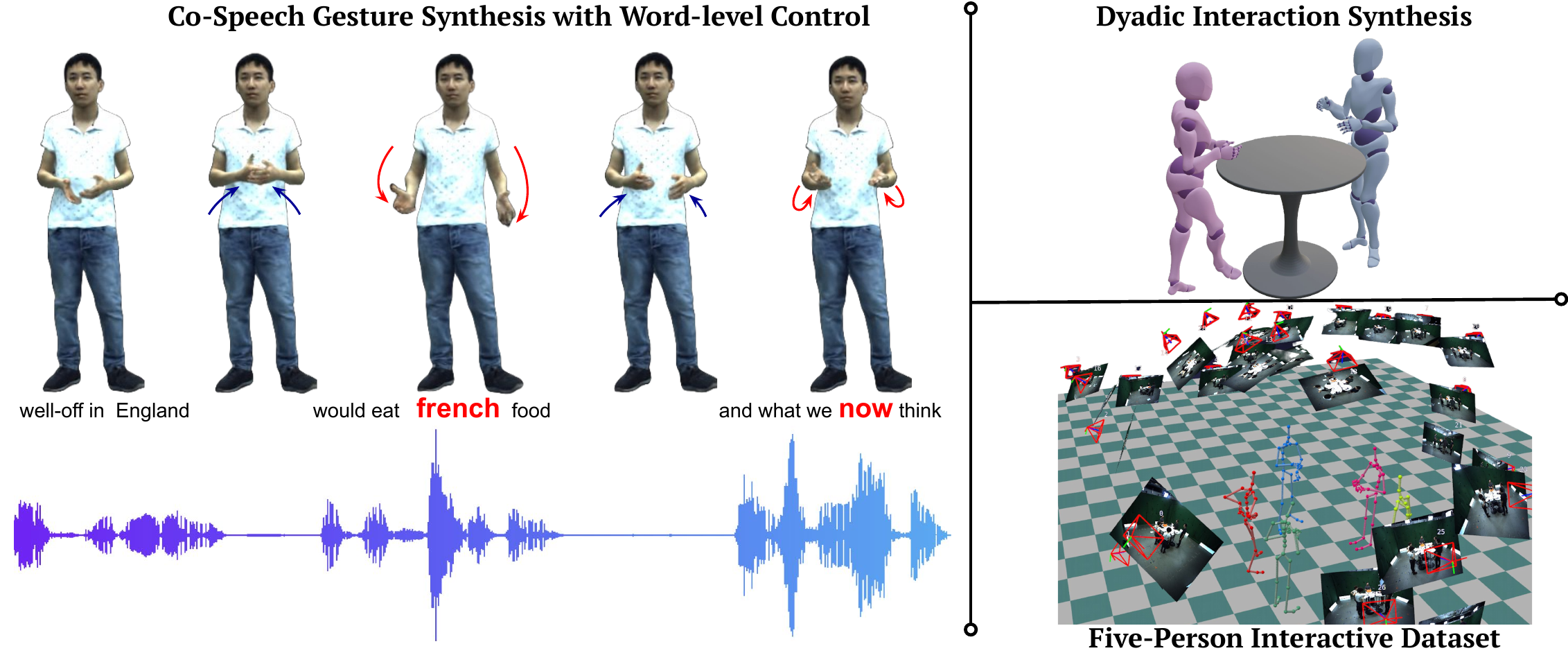}
    \vspace{-15pt}
    \captionof{figure}{\textbf{
    Our \model approach generates body and hand gestures in monadic and dyadic settings, while also offering advanced control over textual and auditory modalities in speech.}  
    \textbf{Lastly, we introduce the \dataset dataset, showcasing rich interactions with co-speech gestures between five participants.}
    Motions rendered using ASH~\cite{zhu2023ash}.
    }
    \label{fig:teaser} 
\end{center} 
}] 
%
%
%
\begin{abstract}
Gestures play a key role in human communication. Recent methods for co-speech gesture generation, while managing to generate beat-aligned motions, struggle generating gestures that are semantically aligned with the utterance. 
Compared to beat gestures that align naturally to the audio signal, semantically coherent gestures require modeling the complex interactions between the language and human motion, and can be controlled by focusing on certain words.
Therefore, we present \model, a diffusion-based approach for multi-modal gesture synthesis,  which can not only generate gestures based on multi-modal speech inputs, but can also facilitate controllability in gesture synthesis. 
Our method proposes two guidance objectives that allow the users to modulate the impact of different conditioning modalities (e.g. audio vs text) as well as to choose certain words to be emphasized during gesturing. 
Our method is versatile in that it can be trained either for generating monologue gestures or even the conversational gestures. 
To further advance the research on multi-party interactive gestures, the \dataset dataset is released, which contains 6 hours of gesture data showing 5 people interacting with one another. 
We compare our method with several recent works and demonstrate effectiveness of our method on a variety of tasks.
We urge the reader to watch our supplementary video at~\href{https://vcai.mpi-inf.mpg.de/projects/ConvoFusion/}{our website}.
\end{abstract}
\vspace{-1.8em}

\section{Introduction}\label{sec:intro} 
Gestures are one of the fundamental ways of expression and
can significantly enhance the interpretation of the verbally communicated utterance~\cite{kendon2004gesture}.
As our society integrates multi-billion parameter large-language-model (LLMs)~\cite{LLMSurvey, touvron2023llama} into our workflows and daily lives, it is only natural to consider ways to augment the LLM based on spoken language alone with \textit{non-verbal} information essential to interpreting such language.
Towards this goal, speech and text-based gesture generation approaches have come a long way from symbolically representing gestures~\cite{cassell94animated, cassell2000embodied} in a rule-based generation framework~\cite{kopp2006towards} to the state-of-the-art methods trained on human motion capture data~\cite{gesturediffuclip, diffgesture, zhao2023diffugesture}. 
\par
Yet, while the majority of methods successfully capture \textit{beat gestures} that are prosodically aligned with speech, they lack language-based control over the gesture generation and therefore, struggle to generate precise \textit{semantic gestures} that contribute to the overall meaning of an utterance.
This can be attributed to the fact that the motion of beat gestures is temporally well-aligned with the speech signals and generally follows a similar spatial pattern for all speakers and content, therefore, it is easier to model using learning techniques.
On the other hand, semantic coherence has a more complex temporal interplay with the words, their meaning and who the individual speaker is.
\par
In this work, we propose \model -- a novel controllable gesture synthesis method to generate not only co-speech gestures, but also reactive (and passive) gestures.
We follow a latent diffusion approach~\cite{stable_diffusion, mld}, which has the benefit of learning a jitter-free motion representation. 
Unlike existing latent diffusion methods~\cite{mld}, we design our motion latents to be time-aware, thus allowing us to learn temporal correlations between motion and speech along with the ability to perform perpetual gesture synthesis.
\par
Our synthesis model supports a variety of input signals (text and audio of the speakers in the conversation) and provides a framework to control them.
To enable controllable multi-modal inference of our model, we introduce a novel classifier-free guidance training strategy. 
More specifically, instead of dropping the entire multi-modal conditioning signal, we show that selectively replacing the modalities with null-vectors facilitates test-time control over each modality.
Finally, \model also allows us to enhance the micro-gestures associated with a particular word, thanks to the fine-grained textual guidance.
Having the test-time modality control and word-level textual guidance provides us the unique ability to have coarse and fine control of the generated motions; a feature missing in existing gesture synthesis works~\cite{gesturediffuclip, habibie21learning, gen_talkshow}. 
\par
One of the goals of our framework is to model the gestures exhibited in a conversational setting.
Unfortunately, most existing datasets only contain monologues, as in the TED~\cite{yoon19robots} and SHOW~\cite{gen_talkshow} datasets.
Even the datasets recorded in conversational setting~\cite{liu2022beat} provide annotations only for one person.
To address this, we introduce the \dataset dataset.
It involves five participants playing multiple sessions of Dungeons and Dragons -- a popular role-playing game.
The dataset comes with high quality full-body motion capture of all the participants, along with multi-channel audio recordings and text transcriptions.
Thanks to around 6 hours of capture, the \dataset dataset allows us to propose a novel approach to generate gestures in a dyadic setting. 
\par
\begin{figure}[!t]
	\includegraphics[width=\linewidth]{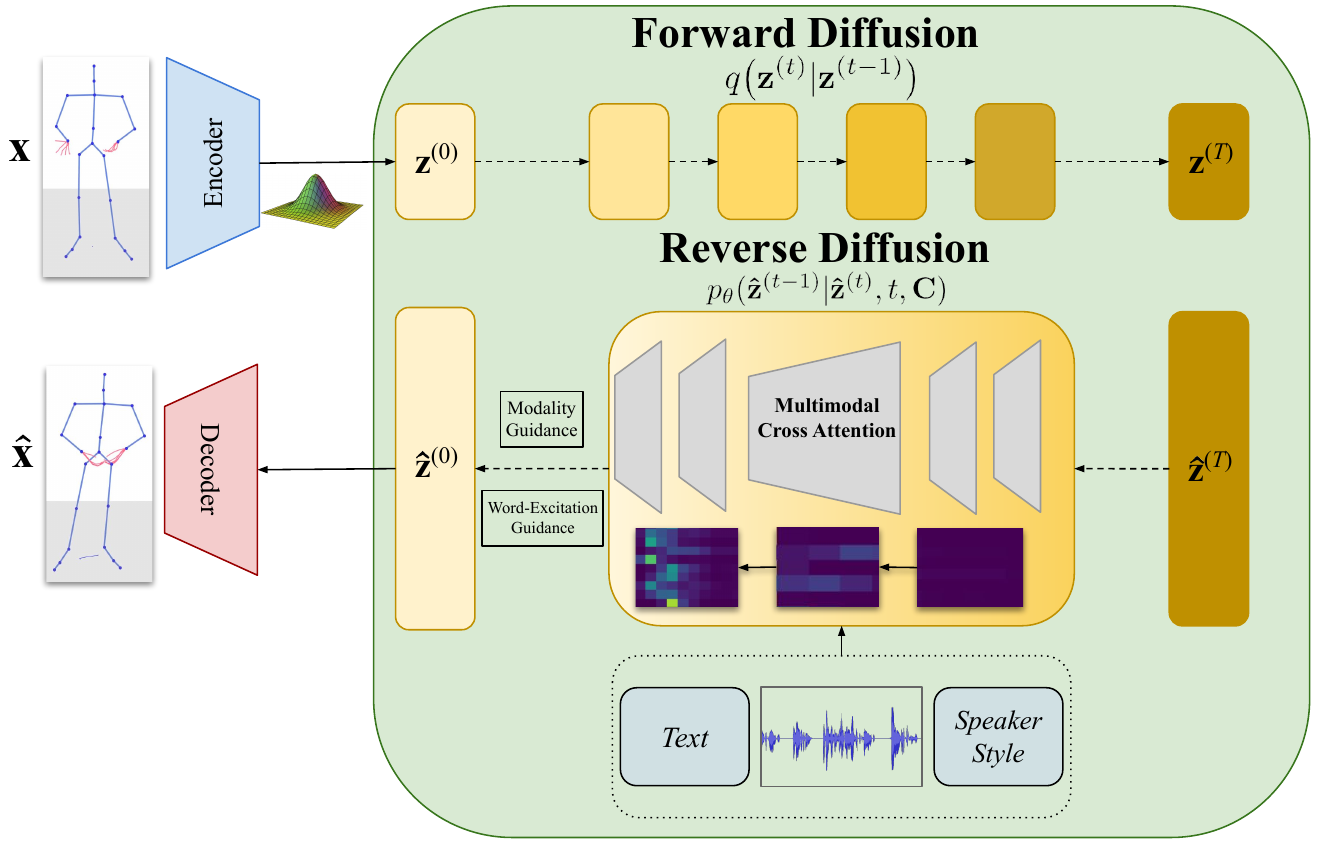}
	\caption{\textbf{Overview of the proposed approach.} We generate gestures conditioned on multiple conditioning signals such as text, audio, speaker style, etc. using a latent diffusion approach. During inference, we introduce modality guidance and word-excitation guidance to control the properties of the generated gestures.}\label{fig:overview}
\end{figure}
In summary, our technical contributions are as follows:
\begin{itemize}
    \item We propose \model -- a diffusion-based approach for monadic and dyadic gesture synthesis.
    We do so not only in the co-speech setting but can also generate passive/reactive gestures.
    \item Thanks to the proposed coarse and fine-grained guidance, our work investigates ways to incorporate a variety of multi-modal signals and provides a framework to control their influence in the generated gestures.
    \item  We demonstrate how generating gestures in the proposed latent mitigates the jittering artifacts prevalent in the hand-articulations of existing datasets.
    Unlike existing motion latent diffusion works~\cite{mld}, the proposed time-aware latent representation allows us to perform perpetual gesture synthesis with high synthesis quality.
    \item This work also introduces the \dataset dataset, thereby facilitating future research on dyadic and group gesture synthesis.
\end{itemize}
\section{Related Works}
\label{sec:related_works} 
As our work draws inspiration from the extensive literature on gesture synthesis and recent works on diffusion-based generative models, we discuss relevant literature from these two perspectives in this section.
\subsection{Co-Speech Gesture Synthesis}
Co-speech gestures are a unique form of gesture, in which hand and arm movements used to communicate information are temporally synchronized and semantically integrated with speech~\cite{marstaller2014multisensory}. While such gestures are thought to contribute to meaning and discourse in the same way as lexical items and intonation patterns, their multi-functional nature makes automatic generation challenging. Non-referential \textit{beat} gestures align with prosodically stressed words and contribute less to overall semantic meaning~\cite{mcneill2008gesture,kendon2004gesture}; such gestures have proved easier to generate~\cite{seeg}. Semantic gestures categorized as \textit{iconic}, \textit{metaphoric}, or \textit{deictic} visually illustrate some aspect of the spoken utterance yet are less patterned between speakers and content; these gestures are more challenging to effectively reproduce \cite{kucherenko2021speech2properties2gestures}.
\par
Early works in the field of co-speech gesture synthesis can be divided into rule-based and data-driven techniques. 
Rule-based methods~\cite{rule1, rule2}, which usually utilize heuristics, generate gesture combinations with high semantic alignment to speech.
\cite{wagner_survey} provides a comprehensive overview of these methods. 
However, they produce unnatural and less diverse gesture outputs.
To mitigate this problem, early statistical approaches~\cite{stat1, stat2} try to model the underlying gesture distribution using data and then predict gestures that are most appropriate for given speech input.
However, both rule-based systems and early statistical approaches predict gesture sequence in terms of known gesticulation units, which makes the final output look unnatural and choppy.
Therefore, recent data-driven learning-based methods~\cite{mlp_based, habibie21learning, rnn_based1, transformer_based, gesturediffuclip} employ neural networks to map speech input to a gesture sequence, which allows for per-frame gesture prediction, providing an end-to-end solution for speech-to-gesture synthesis. 
\cite{simba_survey} provides an in-depth overview of classical and recent data-driven methods.
\par 
Earlier deep-learning-based methods which used CNN~\cite{cnn_based},  RNNs~\cite{rnn_based2, rnn_based3, rnn_based4} and transformers~\cite{transformer_based} employed deterministic approaches to predict gestures for the speech input.
On the other hand, generative methods offer a better alternative since they can introduce stochasticity in the generation process which leads to diverse outputs.
Generative modeling approaches~\cite{gen_gan, gen_vae, gen_vqvae1, gen_flow1, gen_talkshow, zeroeggs} have been used for synthesis resulting in human-like gestures.
But, they also suffer from low semantic relation with the speech input because there exists many-to-many relations between speech and gestures and it becomes hard for the generative approaches to realize which gesture is more semantically accurate corresponding to the speech.
Therefore, recent approaches~\cite{seeg, semantic1, rhythm_gest, gesturediffuclip, disco} try to improve intent's alignment with gesture prediction. 
Gesture styles are also incorporated in the gesture generation pipeline for personalized gesture synthesis~\cite{diffstylegest, zeroeggs}.
\subsection{Speech Gesture Datasets}
As the performance of learning-based methods relies on the quality of its training data, a number of gesture synthesis datasets have been proposed by the community. 
However, high-quality speech-driven gesture synthesis datasets are typically expensive and tedious to collect as they require hours of speech gesture motion capture (mocap) recordings in a studio setting. 
Because of these limitations, early works typically involve a single speaker \cite{ferstl18investigating, ferstl2020adversarial}. 
To collect a large number of training samples, several works have proposed to leverage monocular 3D estimation approaches to obtain the 3D body, face, and hand keypoints \cite{ginosar19learning, habibie21learning, ahuja2020style, yoon19robots, gen_talkshow, ghorbani2022zeroeggs}. 
Unfortunately, such monocular estimation results are subpar compared to the standard multi-view mocap approaches and are unsuitable for multi-speaker settings.
\par
To address the lack of large-high-quality data,~\cite{liu2022beat} proposed BEAT, a 76-hour mocap-based speech gesture dataset recorded from 30 different subjects. 
Unlike BEAT which focused on a single speaker,~\cite{lee2019talking} introduced a high-quality speech gesture dataset that involved multiple speakers, but was limited to two-person conversations. 
In contrast to previous works, we propose a high-quality speech-gesture dataset involving 5 subjects within a conversation. In addition, different from most mocap-based datasets that use marker-based mocap technologies, we employ a state-of-the-art markerless mocap system to accurately capture the 3D body and hands of multiple speakers without being restricted by body mocap suits.
\cref{tab:dataset_tab} provides a brief overview of some notable datasets and their qualities.
Moreover, we also compare them with the \dataset dataset we present in this work. 
\begin{figure*}
    \centering
    \includegraphics[width=0.95\textwidth]{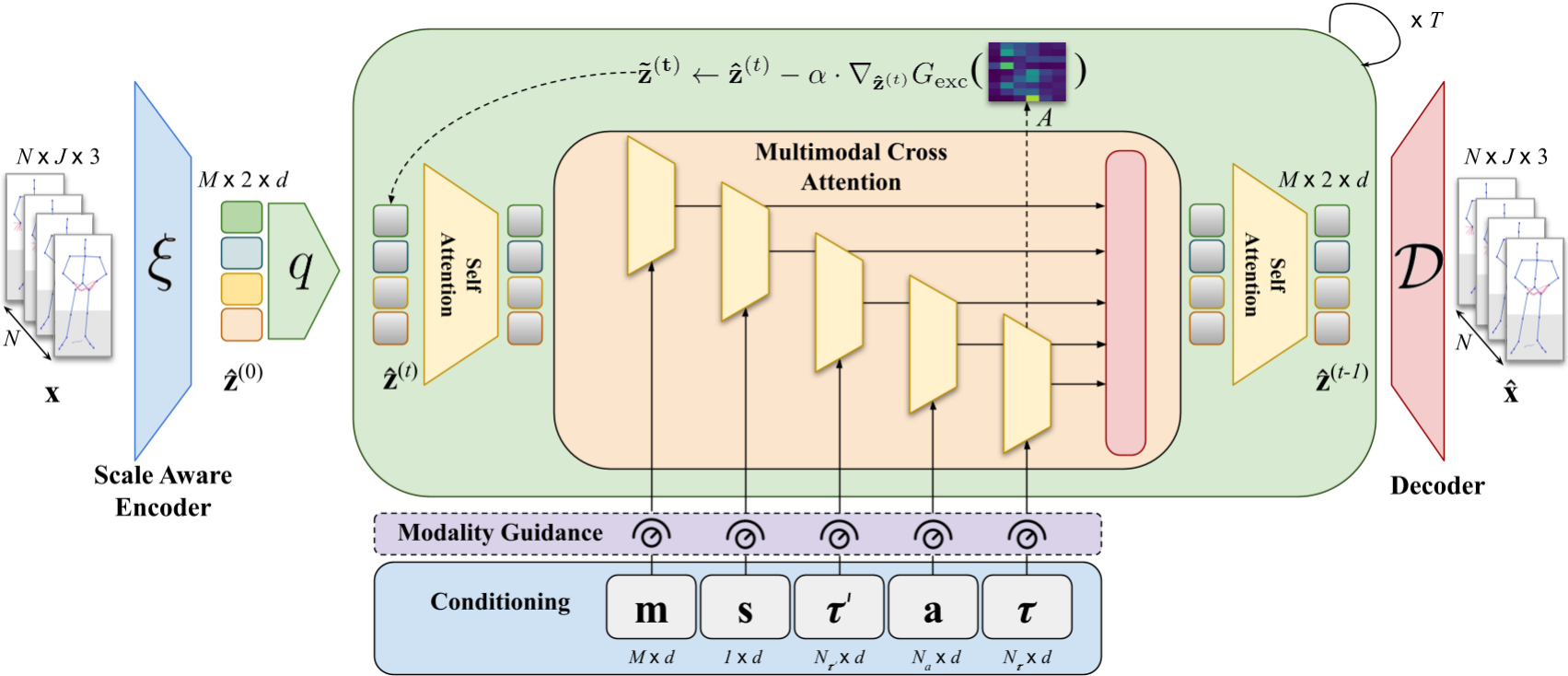}
    \caption{\textbf{The model schema.}
    Given a training motion $\mathbf{x} \in \mathbb{R}^{N\times J \times 3}$, we first extract its latent encoding $\mathbf{\hat{z}}^{(0)}$ (\cref{ssec-m:uncond}), which is then denoised by a network that incorporates the various modalities in the denoising process.
    At inference time, the denoised latents are decoded to produce the final generation, $\mathbf{\hat{x}}$ (\cref{ssec-m:cond-gesture-gen}).
    During this process, our method allows to control the generation through coarse-grained modality guidance or fine-grained word-excitation guidance (\cref{ssec-m:guidance-control}).
    Dotted lines represent components used only during inference.
    }
    \label{fig:main-arch}
    \vspace{-1.0em}
\end{figure*}
\subsection{Diffusion-based Generative Modelling}
Diffusion models \cite{sohldick, ddpm} have demonstrated remarkable potential in the field of generative modeling, consistently delivering impressive results in various synthesis applications~\cite{dalle2, imagen, diffwave, mofusion, edge, physdiff, socialdiffusion}.
%
%
%
New paradigms like guidance mechanisms~\cite{diffusion-beat-gan, clf-guidance} and latent diffusion models~\cite{ldm} have been introduced to enhance quality and alignment of diffusion-based synthesis w.r.t given conditionings.
\par
This approach has been extensively applied for conditional human motion synthesis~\cite{mdm, mofusion, edge, mld}.
Similarly, co-speech gesture generation has also greatly benefited from this generative modeling technique. 
DiffGesture~\cite{diffgesture} uses a transformer-based diffusion pipeline with an annealed noise sampling strategy for temporally consistent gesture generation.
GestureDiffuCLIP~\cite{gesturediffuclip} employs latent-diffusion models~\cite{ldm} and CLIP~\cite{clip} based conditioning to improve control over co-speech gesture generation.
\cite{socialdiffusion} presents a model to predict the movement of multiple speakers in a social setting.
However, contrary to other diffusion-based gesture synthesis approaches, their model focuses on predicting the correctness of the 3D body keypoint trajectory for a few seconds in the future instead of improving the speech-gesture alignment.
Instead of simply predicting the motion trajectory, our method proposes a multi-person speech-driven 3D gesture synthesis approach that can be used to predict the 3D reactive body and hand motion between various speakers and listeners within a conversation. 
\section{Approach}
\label{sec:method} 
The goal of our method is to generate co-speech gesture sequences for monadic and dyadic settings in correspondence with input speech.
A gesture sequence $\mathbf{x} \in \mathbb{R}^{N\times J \times 3}$ consists of $N$ frames of human motion with $J$ articulating 3D joints.
The generated gesture motion ought to be consistent with the multi-modal conditioning signal, $\mathbf{C}$, representing the speech and identity-related attributes of the persons in conversation (discussed later in~\cref{ssec-m:cond-gesture-gen}).
\par
We design our gesture synthesis method around a latent denoising diffusion probabilistic model (DDPM) framework~\cite{stable_diffusion}.
The proposed diffusion model is trained to denoise the latent representation of the gesture motions (refer to~\cref{ssec-m:uncond}).
The generated motion latents can later be decoded using a motion decoder.
Unlike existing motion latent diffusion methods~\cite{mld}, we design our latent space in a time-decomposable manner, thereby allowing us to learn fine-grained interplay between motion and speech.
Crucially, our method also allows the end-user to \textit{control} the attributes of the generated gestures at inference time (see~\cref{ssec-m:guidance-control}).
We now discuss each component in detail. 
Refer to the supplemental document for a glossary of major notations used in the method explanation.
\par
\subsection{Scale-aware Temporal Latent Representation}
\label{ssec-m:uncond}
Instead of directly denoising the raw motion $\mathbf{x}$, our diffusion model operates in the latent space of human motion.
Thus, we propose to learn such a latent space with two characteristics:
1) We disentangle the finger motions from the rest of body motions by encoding them into a latent space through separate encoders.
2) Instead of projecting the entire motion into one single latent vector, we encode motion into chunked latents that can be decoded jointly by a decoder.
\par
\noindent
\textbf{Decoupled Latent Representations.} The articulation of the finger joints is critical to the quality of gesture synthesis.
However, the fingers articulate in a significantly different space and scale compared to the rest of the body and na\"ively encoding the full-body gestures results in inaccurate reconstruction of hands.
We therefore follow prior works that decouple the two sets of joints~\cite{ghosh2022imos, ghosh_2021} and represent the motion $\mathbf{x}$ as a latent vector $\mathbf{z} = \{\mathbf{z}_b, \mathbf{z}_h\}$, where $\mathbf{z}_b \in \mathbb{R}^{d}$ and $\mathbf{z}_h \in \mathbb{R}^{d}$ are separate encodings of the body and hand motion.
\par
The latent vectors are learned using a VAE framework.
The hand and body motions, $\mathbf{x}_h$ and $\mathbf{x}_b$, are encoded using transformer encoders: $\mathbf{z}_b = \xi_b(\mathbf{x}_b),~\mathbf{z}_h = \xi_h(\mathbf{x}_h)$.
The latent vectors represent the mean of the distribution, which can be sampled using the reparameterization trick~\cite{vae} and fed into a decoder to reconstruct the motion 
${\mathbf{x}}_b^{\prime} = \mathcal{D}_b(\mathbf{z}_b),{\mathbf{x}^{\prime}}_h = \mathcal{D}_h(\mathbf{z}_h)$.
We train the VAE with the standard reconstruction loss, $\mathcal{L}_2$, Bone-length regularization loss $\mathcal{L}_{bone}$~\cite{mofusion} and the KL-Divergence of the latents, $\mathcal{L}_{KL}$.
Additionally, we reduce the jitter in reconstruction proposing a Laplacian regularization term:
\begin{equation} \label{eq:laplacian}
   \mathcal{L}_{lap} = \norm{\mathscr{L}\{\hat{\mathbf{x}}\} - \mathscr{L}\{\mathbf{x}\}}_2
\end{equation}
where $\mathscr{L}\{\cdot\}$ is the Laplace transform operator along $N$ frames.
Refer to~\cref{ssec:ablations} and supplemental for analysis.
%
\par
\begin{figure}[h]
	\includegraphics[width=0.95\columnwidth]{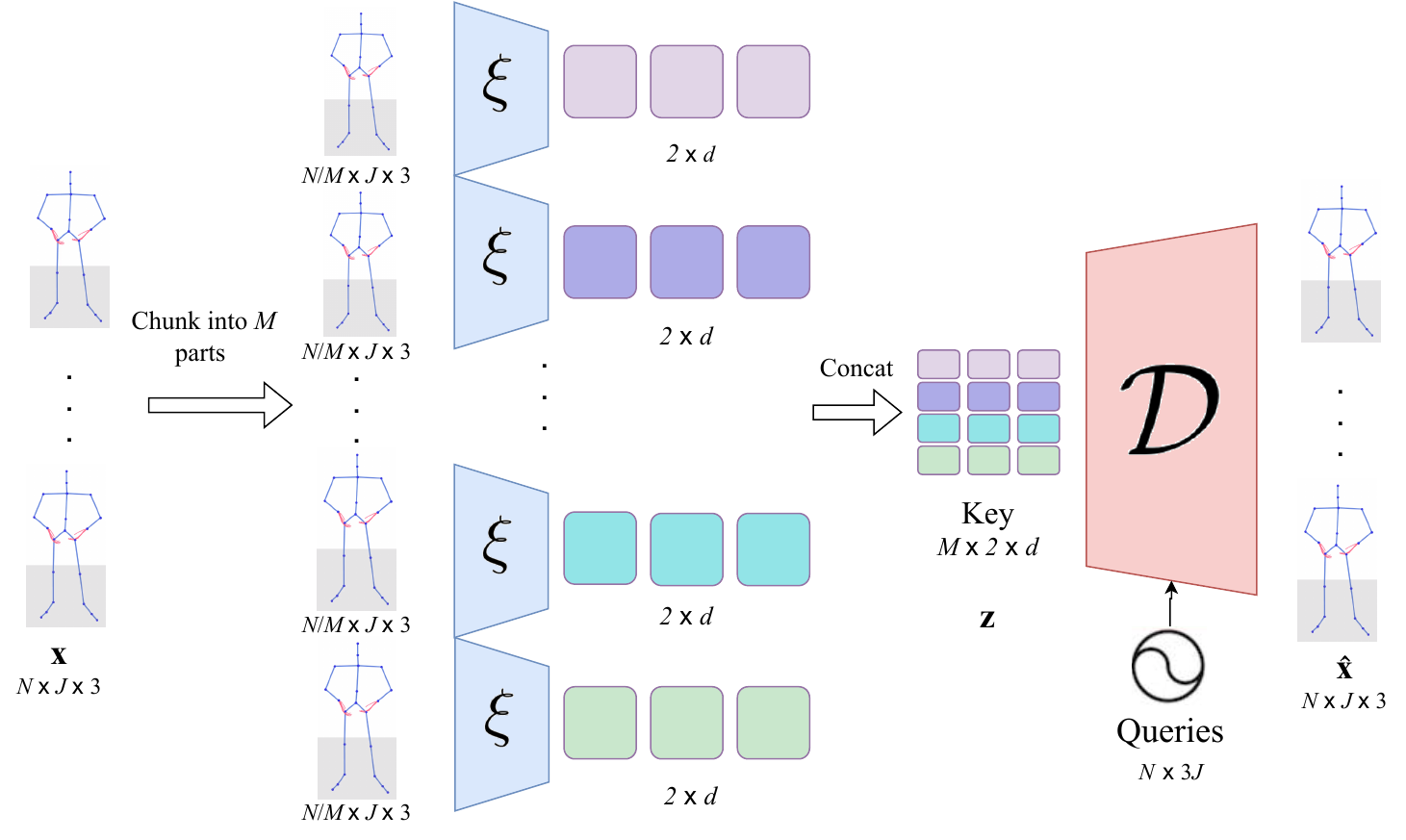}
	\caption{\textbf{Chunked latent encoding-decoding.} We encode a motion of $N$ frames into a sequence of $M$ latent vectors, which are jointly decoded by the decoder $\mathcal{D}$. Encoding into chunked latents allows for perpetual rollout and decoding jointly induces temporal consistency while converting the latents back into motion.}
	\label{fig:timeaware}
    \vspace{-1.0em}
\end{figure}

\noindent
\textbf{Time-Aware Latent Representation.}
The motion latents learned by the VAE represent a large motion sequence ($>$100 frames) with a single $d$-dimensional vector.
This, rather coarse, granularity prohibits applications such as perpetual rollout where the motion can be autoregressively decoded with an overlapping window.
To enable such applications, we propose to encode shorter motion chunks in the latent $\mathbf{z}$ but decode multiple such chunked latents, $\{\mathbf{\hat{z}}_i\}_{i=1}^M$ together with a single decoder, as shown in~\cref{fig:timeaware}.
\par
Given a gesture sequence $\mathbf{x}$, we first split the sequence into $M$ equally sized chunks 
$\{\mathbf{{x}}_i^{\prime}\}_{i=1}^{M}$, where $\mathbf{{x}}_i^{\prime} \in \mathbb{R}^{N / M \times J \times 3}$. 
Next, each of the chunks $\mathbf{{x}}_i^{\prime}$ is encoded in isolation using $\mathbf{\hat{z}}_i = \xi_b(\mathbf{{x}}_i^{\prime})$.
However, while decoding, the decoder collectively decodes a sequence of chunked latents, $\mathbf{\hat{z}} = \{\mathbf{\hat{z}}_i\}_{i=1}^M$, following: $\mathbf{\hat{x}}^{\prime} = \mathcal{D}(\mathbf{\hat{z}})$.
In summary, our latent encodings transform a motion sequence $\mathbf{x} \in \mathbb{R}^{N \times J \times 3}$ into latent representations $\mathbf{\hat{z}} \in \mathbb{R}^{M \times 2 \times d}$.
This can enable perpetual gesture generation using diffusion inpainting technique~\cite{lugmayr2022diffinpaint} as we discuss and analyze in~\cref{ssec:ablations}.
%
%
\subsection{Modality-Conditional Gesture Generation}
\label{ssec-m:cond-gesture-gen}
Having obtained $\mathbf{\hat{z}}$ as the time-aware latent representation of gesture motions, we formulate the gesture synthesis task as that of conditional latent diffusion~\cite{stable_diffusion}.
%
%
%
The forward diffusion process, successively corrupts the latent sequence $\mathbf{\hat{z}}^{(0)}$ by adding Gaussian noise $\epsilon$ for $T$ timesteps with the assumption that $\mathbf{\hat{z}}^{(T)} \sim \mathcal{N}(0, I)$.
For generation, the \textit{reverse diffusion} process is performed on $\mathbf{\hat{z}}^{(T)}$ by iteratively denoising $\mathbf{\hat{z}}^{(T)} \sim \mathcal{N}(0, \mathbf{I})$ to generate a latent sequence $\mathbf{\hat{z}}^{(0)}$, and can be formulated as
\begin{equation}
p_\theta\big(\mathbf{\hat{z}}^{(0:T)}\big) = p\big(\mathbf{\hat{z}}^{(T)}\big) \prod_{t=1}^T p_\theta\big(\mathbf{\hat{z}}^{(t-1)} | \mathbf{\hat{z}}^{(t)}\big),    
\vspace{-0.7em}
\end{equation}

where $p_\theta(\mathbf{\hat{z}}^{(t-1)} |\mathbf{\hat{z}}^{(t)})$ is approximated using a neural network parameterized by weights $\theta$.
This neural network $f_\theta$ is trained to predict noise $\epsilon_{\theta}(\mathbf{\hat{z}}^{(t)}, t)$ \cite{ddpm}, which can be used in the training objective $\mathcal{L}_d = || \epsilon - \epsilon_{\theta}(\mathbf{\hat{z}}^{(t)}, t) ||^2.$
\par
The motion generation framework discussed above is so far \textit{unconditional}.
Our gesture synthesis approach can be conditioned in primarily two settings: monadic and dyadic.
The \textit{monadic setting} refers to the co-speech gesture generation based solely on the speaker's own utterance and typically occurs in monologue scenarios.
For this, we represent the conditioning signal as $\mathbf{C} = \{\mathbf{a}, \boldsymbol{\tau}, \mathbf{s}\}$, consisting of the audio signal $\mathbf{a} \in \mathbb{R}^{N_a \times d}$ and the text tokens $\boldsymbol{\tau} \in \mathbb{R}^{N_{\boldsymbol{\tau}} \times d}$, as well as $\mathbf{s} \in \mathbb{R}^{1 \times d}$ representing the speaker identity token.
Generally, $N_a$ corresponds to the number of audio frames, $N_{\boldsymbol{\tau}}$ corresponds to the number of text tokens in the utterance.
Speaker identity $\mathbf{s}$ can enable applications like stylized gesture synthesis which can generalize to different gesture styles.
For the \textit{dyadic setting}---which takes place in conversation scenarios---the generated gestures must be in accordance with the co-participant's utterance as well.
In this case, we have $\mathbf{C} = \{\mathbf{a}, \boldsymbol{\tau}, \boldsymbol{\tau}^{\prime}, \mathbf{s}, \mathbf{m}\}$, where $\boldsymbol{\tau}^{\prime}$ refers to the co-participant's speech content i.e. their text.
Here, we can also choose their audio instead of their text as well.
Finally, $\mathbf{m} \in \{0, 1\}^{M}$ indicates whether the speaker is actively responding with speech, or passively back-channeling e.g. by laughing or nodding (see also supplemental video).
\par
We use a transformer decoder network~\cite{vaswani2017posenc} with multi-head attention to approximate the denoising function producing $\epsilon_{\theta}(\mathbf{\hat{z}}^{(t)}, t, \mathbf{C})$.
This allows us to elegantly integrate multiple modalities in $\mathbf{C}$ with separate cross-attention heads, as shown in~\cref{fig:overview}.
Let us consider the case of the audio signal, $\mathbf{a}$.
The cross-attention features, $\phi_a$, are computed using the attention matrix $\operatorname{Attn}(\mathbf{\hat{z}}, \mathbf{a}) \in \mathbb{R}^{N_a \times M}$ as:
\begin{equation}
    \operatorname{Attn}(\mathbf{\hat{z}}, \mathbf{a}) = \sigma(\frac{Q_z K_a}{\sqrt{d}}),
    \phi_a = V_a \cdot \operatorname{Attn}(\mathbf{\hat{z}}, \mathbf{a})
\end{equation}
where $\sigma$ is the softmax operator, $Q_z, K_a, V_a$  are the query, key and value vectors recovered form the motion latent features $\mathbf{\hat{z}}$ and the audio features $\mathbf{a}$.
We similarly recover text features, $\phi_{\tau} = \operatorname{Attn}(\mathbf{\hat{z}}, \boldsymbol{\tau})$, also for the text.
\par
\subsection{Towards Controllable Gesture Generation}
\label{ssec-m:guidance-control}
In addition to multi-modal gesture synthesis, our method is designed to allow coarse and fine-grained control.
For coarse control, one can adjust the impact of a specific modality on the generated motion by utilizing our \textit{modality-level guidance strategy}.
For fine control, the user can choose specific words to enhance the gestures for the words using the proposed \textit{word-excitation guidance} (WEG) objective.
\par
\noindent
\textbf{Modality-Guidance.}
Classifier-free guidance~\cite{clf-guidance} has been used to improve the generation quality of various diffusion-based motion and gesture generation methods~\cite{gesturediffuclip, mld, mdm, kulkarni2023nifty}.
Typically, this is done by randomly replacing the conditioning vectors with a null-embedding $\mathbf{C} \leftarrow \emptyset$.
At inference, the noise predictions are blended at each diffusion timestamp $t$ to get the noise prediction $\epsilon_{\theta}^{(t)}$:
\begin{equation}
    \epsilon_{\theta}^{(t)} = \epsilon_{\theta}(\mathbf{\hat{z}}^{(t)}, t, \emptyset) + \lambda (\epsilon_{\theta}(\mathbf{\hat{z}}^{(t)}, t, \mathbf{C}) - \epsilon_{\theta}(\mathbf{\hat{z}}^{(t)}, t, \emptyset)) 
\end{equation}
where, $\lambda$ represents the guidance scale.
Once estimated, $\epsilon_{\theta}^{(t)}$ can be used to sample $\mathbf{\hat{z}}^{(t-1)}$ for the next iteration using Eq. 11 of~\cite{ddpm}.
However, recall that our conditioning set $\mathbf{C} = \{\mathbf{a}, \boldsymbol{\tau}, \boldsymbol{\tau}^{\prime}, \mathbf{s}, \mathbf{m}\}$ consists of several modality-specific conditions.
Na\"ively setting all the elements to $\emptyset$ for random iterations prohibits separately learning the effect of each individual modality within $\mathbf{C}$ on the conditional distribution.
Instead, we train our model with random modality dropouts (with null-embedding replacement) with $10\%$ drop probability.
This encourages the model to learn several combinations of marginalized conditional probability distributions.
\par
At inference, we sample with modality-guidance:
\begin{gather} \label{eq:modality_guidance}
        \epsilon_{\theta}^{(t)} = \epsilon_{\theta}^{\emptyset} +
    \lambda_m \sum_{\mathbf{c} \in \mathbf{C}} w_c \big( \epsilon_{\theta}^{c}(\mathbf{\hat{z}}^{(t)}, t, \mathbf{c}) - \epsilon_{\theta}(\mathbf{\hat{z}}^{(t)}, t, \emptyset) \big)
\end{gather}
where the scale parameters, $w_{c} \geq 0$, determine the contribution of each modality towards the generated gesture and $\lambda_m$ is the global guidance scale.
Adjusting the modality scale, $w_{c}$ allows us to coarsely control the gesture quality and also analyze the sensitivity of the generation process to specific modalities.
Note, that this is an optional sampling strategy required only for modality-level control.
\par
\noindent
\textbf{Word-Excitation Guidance.}
Inspired by the controllable image generation methods~\cite{chefer2023attendexcite, epstein2023selfguidance}, we propose a word-level guidance mechanism that allows us to finely control the gesture generation based on a user-defined set of words during the sampling process.
\par
Let $\operatorname{Attn}(\mathbf{\hat{z}}^{(t)}, t, \boldsymbol{\tau}) \in \mathbb{R}^{N_{\boldsymbol{\tau}} \times M}$ be the text attention matrix at the $t^{th}$ iteration of the denoising process.
For a set of text tokens $\{\boldsymbol{\tau}_i\}_{i=1}^{S}$, selected by a \textit{user} with the intention of gesture enhancement, we focus on the corresponding column, $A_{i} \in \mathbb{R}^M$ in the text attention matrix.
Now, with the assumption that the element with maximum attention in $A_i$ aligns with the motion chunk associated with the text, we introduce a guidance objective to further enhance (or, excite) the same attention: 
\begin{equation}
    G_{\mathrm{exc}} = \frac{1}{S} \sum_{i=1}^{S} (1 - \operatorname{max}(A_i))
    \vspace{-0.5em}
\end{equation}
Next, we use the gradient of $G_{\mathrm{exc}}$ w.r.t the latent $\mathbf{\hat{z}}^{(t)}$ to perform the word-excitation guidance:
 \begin{equation}
    \mathbf{\Tilde{z}}^{(t)} \leftarrow \mathbf{\hat{z}}^{(t)} - \alpha \cdot \nabla_{\mathbf{\hat{z}}^{(t)}}G_{\mathrm{exc}},
     \epsilon_{\theta}^{(t)} = f_{\theta}(\mathbf{\Tilde{z}}^{(t)}, t, \mathbf{C})
\end{equation}
where $\alpha$ is the guidance scale for the word excitation guidance, which also serves as a step size for latent update.

\section{Dataset}
To enable a high-quality, speech-driven gesture synthesis method involving multiple speakers, we introduce the \dataset dataset. 
Our dataset is designed to also invoke a wide range of non-verbal gestures during the speaker interactions. We based our dataset recordings on D\&D tabletop roleplaying game, where five different players are standing in a circle around a game map. Each participant is equipped with a dedicated wireless microphone to ensure a clean audio recording and audio source separation. 
The setup of the gameplay involves various types of interaction between the actors that often require semantically meaningful gestures such as pointing to a certain location on the map.
In total, the dataset consists of 4 separate recording sessions with a total duration of 6 hours. 

Our proposed dataset is recorded using a state-of-the-art multi-view markerless mocap to obtain accurate 3D body and hand pose estimates of multiple subjects at a given time. This allows our participants to move freely without being obstructed by the tight mocap suit or gloves. In addition to audio and the 3D pose annotations, we also provide text and gesture annotations for each individual speaker that distinguishes different types of observable gestures, including beats, iconic, deictic, and metaphoric. Our dataset will be made publicly available to the community.
\begin{table}
    \centering
    \resizebox{\columnwidth}{!}{
    \begin{tabular}{|l|c|c|c|c|c|} \hline  
         Name&  \# Identities &Size&Body Parts& \shortstack{Multi-party\\ Interaction} & \shortstack{\# Interacting\\ Speakers} \\ \hline 
         IEMOCAP~\cite{busso2008iemocap}& 10& 12h& Face& \Checkmark&2\\ \hline 
         Creative-IT~\cite{metallinou2016usc}& 16& 2h& Body $\dagger$& \Checkmark&2\\\hline
         CMU Haggling Dataset~\cite{Joo_2017_TPAMI}&  122&3h&Face, Body, Hands& \Checkmark&3 \\ \hline  
         TED Dataset~\cite{yoon19robots} &   1295&52.7h&Upper Body& & \\ \hline  
         Speech Gesture 3D~\cite{habibie21learning} &   10&  144h& Upper Body, Hands, Face& & \\ \hline
         Talking with Hands~\cite{lee2019talking}&   50&50h&Body, Hands& \Checkmark& 2\\ \hline 
         PATS~\cite{ahuja2020style}& 25& 250h& Upper Body, Hands& &\\\hline 
         SaGA++~\cite{kucherenko2021speech2properties2gestures}& 25& 4h& Body hands& & \\ \hline 
         ZeroEGGS Dataset~\cite{ghorbani2022zeroeggs}& 1& 2h& Body, Hands& & \\ \hline 
         BEAT~\cite{liu2022beat}& 30& 76h& Body, Hands, Face& & \\\hline \hline  
         \dataset &  5 &6h&Body, Hands& \Checkmark&5 \\ \hline 
    \end{tabular}}
    \caption{Comparison of currently available datasets to our \dataset dataset. 
    \textit{Body parts} refer to the parts where the 2D or 3D pose tracking is available. 
    $\dagger$ indicates that the body tracking is only available for one of one interacting actors. 
    }
    \label{tab:dataset_tab}
\end{table}
\section{Experiments} \label{sec:experiments}
Our method, in its vanilla form, is designed to generate human gestures from speech, yet it goes several steps beyond this task.
For instance, we adapt our method to perform dyadic conversations. 
More importantly, we show how different modalities contribute to the generation and perform fine-grained text-based control.
Naturally, it is difficult to find suitable baselines to compare with.
To perform fair evaluations, we, therefore, compare with methods that can be trivially adapted to our setting.
Specifically, we compare with MLD~\cite{mld} (a generic latent diffusion method), CaMN~\cite{liu2022beat}, Multi-Context~\cite{rnn_based3}, DiffGesture~\cite{diffgesture} (specifically monadic gesture baselines) and DiffuGesture~\cite{diffugesture} (two-person motion synthesis works).
Notably, CaMN~\cite{liu2022beat}, DiffGesture~\cite{diffgesture} and DiffuGesture~\cite{diffugesture} require a seed motion sequence to build the gesture generation on.
This is different from our setting and provides vital clues about the gesture style.
We provide the seed motions for the two methods, but do not use the seed motions to generate our results.
The methods are compared using the established motion synthesis metrics as well as a user-study.
\par
\noindent \textbf{Evaluation Datasets.} 
We evaluate our performance in monadic gesture generation on the recently introduced BEAT dataset~\cite{liu2022beat}. 
The test set includes $2492$ $5$-sec motion sequences and includes a set of $5$ unseen speakers.
For evaluating the motion in dyadic setting, we use the test set of the proposed \dataset dataset.
The test set contains $3932$ sequences of $5$-second conversations.

\par
\noindent \textbf{Metrics.} Evaluating synthesized motions is challenging due to the subjective nature of perceiving good gestures.
Yet, we evaluate our method on the established metrics like Beat-Alignment~\cite{siyao2022bailando}, FID, Semantic Relevance Gesture Recall (SRGR)~\cite{liu2022beat} that evaluate different aspects of the motion.
We also use Diversity and L1 Divergence to evaluate the ability of models to span the space of gesture motions with enough coverage.
\begin{table}
    \centering
    \resizebox{\columnwidth}{!}{
    \begin{tabular}{|c|c|c|c|c|c|} \hline 
         &  FID $\downarrow$&  BeatAlign $\rightarrow$&  Diversity $\rightarrow$& L1 Div $\rightarrow$& SRGR $\uparrow$\\
         \hline
         GT& -& 0.89& 13.21& 13.12&-\\\hline \hline 
         Multi-Context~\cite{rnn_based3}&  $\geq10^3$ &  $0.8$ &  $26.71$ &  $43.31$ & $0.140$\\ \hline 
         DiffGesture~\cite{diffgesture}&  $\geq10^3$ &  $0.96$ &  $176$&  $17.8$ & $0.003$\\ \hline  \hline
         CaMN \cite{liu2022beat}& \gold{$142$} &  $0.74$ &  \silver{$9.66$}& \silver{$5.85$} & \gold{$0.443$}\\ \hline 
         MLD~\cite{mld}&  $475$ & \silver{$0.76$} & $16.98$ &  $5.42$&$0.214$\\ \hline 
         Ours&  \silver{$271$} & \gold{$0.82$} &  \gold{$9.82$} & \gold{$6.24$} & \silver{$0.365$}\\ \hline
    \end{tabular}
    }
    \caption{\textbf{Comparison on the BEAT~\cite{liu2022beat}.} Two methods~\cite{rnn_based3, diffgesture} produce extremely jittery motions. We demonstrate superior beat alignment and diversity scores among the remaining methods.}
    \label{tab:monadic_beat}
\end{table}
\subsection{Monadic Co-speech Gesture Synthesis}
We tabulate our results on the BEAT test set for monadic co-speech gesture synthesis in~\cref{tab:monadic_beat}.
We observe that DiffGesture~\cite{diffgesture} and Multi-ContextNet~\cite{rnn_based3} struggle with the FID which, upon visualization, can be attributed to the extremely jittery nature of the generated motions.
Interestingly, this also leads to Multi-ContextNet~\cite{rnn_based3} to perform the best in the Beat Alignment metric as for every beat in the audio, there is always a jittery motion to align with.
Among other methods, we observe better performance in terms of diversity and beat alignment.
It is interesting to note that MLD, which is trained on a non-temporal latent representation, achieves a reasonable beat alignment but worse semantic recall.
We hypothesize that the semantic alignment benefits from a finely discretized motion representation.
Our method lies in the middle of the discretization spectrum, where CaMN operates on raw motion frames while MLD collapses the temporal axis within a single latent.
\subsection{Dyadic Co-speech Gesture Synthesis}
We adapted two baselines to the dyadic setting for comparison.
MLD's architecture was extended by adding additional conditioning blocks of the co-participant's speech.
\begin{wraptable}{l}{0.6\columnwidth}
    \resizebox{0.6\columnwidth}{!}{
    \begin{tabular}{|c|c|c|c|} \hline 
         &  BeatAlign $\rightarrow$ &  Diversity $\rightarrow$ & L1 Div $\rightarrow$\\ \hline
 GT& $0.90$ & $17.7$& $5.12$\\\hline\hline 
         MLD& $0.96$ & \gold{$20$}&$0.31$\\ \hline 
 DiffGesture& $0.97$& $2176$&$1308$\\\hline \hline 
         Ours& \gold{$0.90$}& $6.38$&\gold{$1.19$}\\ \hline
    \end{tabular}}
    \caption{\textbf{Qualitative comparison of dyadic motion synthesis on the \dataset dataset.}}
    \label{tab:dyadic_dnd}
    \vspace{-1.0em}
\end{wraptable}
Likewise, DiffuGesture was adapted to our setting as detailed in~\cite{diffugesture}.
We observe similar patterns of jittery motion with DiffuGesture, whereas MLD produced suboptimal results in terms of beat alignment. 
In contrast, we achieve similar beat alignment as the ground-truth while also producing higher L1 Diversity, thus indicating non-static motions. 
\par
\begin{figure}
    \centering
    \includegraphics[width=\linewidth]{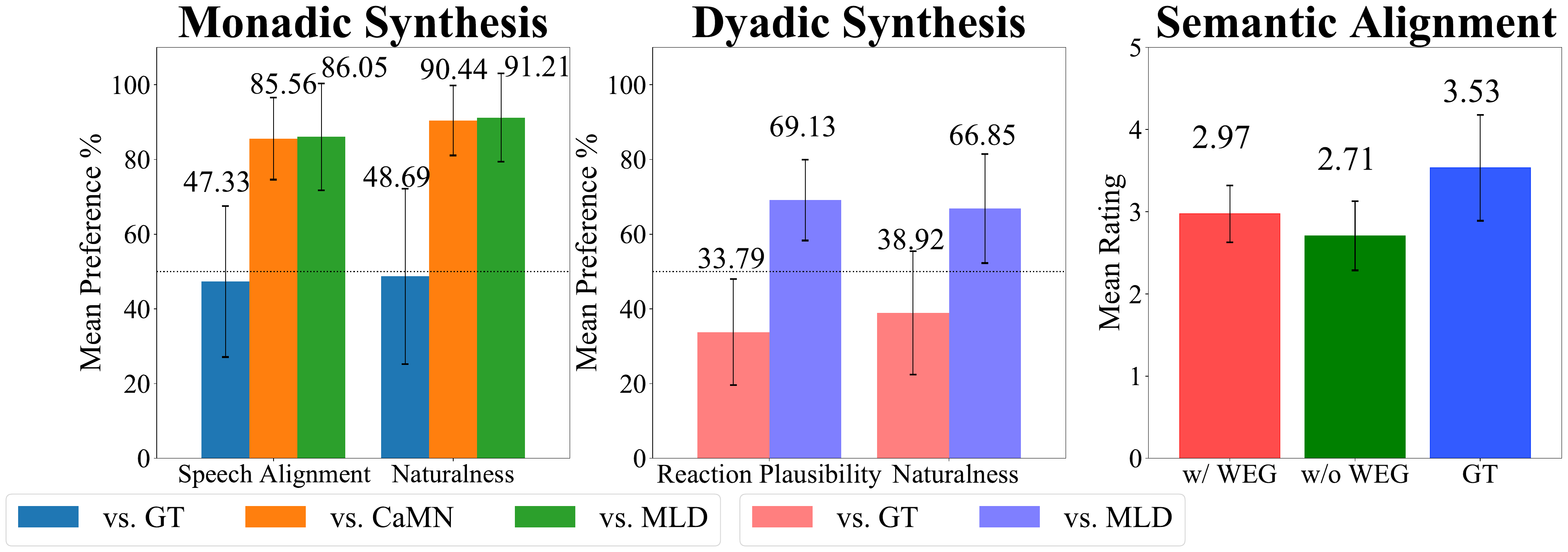}
    \caption{\textbf{Results of the user study.} We compare with CaMN~\cite{liu2022beat} and MLD~\cite{mld}, and achieve an overall favourable preference scores for monadic and dyadic settings. We also evaluate the effectiveness of the word-excitation guidance (WEG).}
    \label{fig:userstudy}
\end{figure}
\subsection{User Study}
As noted above, evaluating motion synthesis models on a set of numerical metrics hides several aspects of the gesture synthesis.
Prior works~\cite{edge, mofusion} report mismatch between metrics and the subjective evaluations by the users.
Hence, we perform a perceptual user study to evaluate the quality of our synthesis results w.r.t state-of-the-art methods.
For evaluating the monadic results, we aim to evaluate the general plausibility of the motions and probe the coherence of the gestures with the utterance.
Likewise for dyadic synthesis, the goal is to measure if participant's generated gestures align well with their speech as well as co-participant's speech content.
To evaluate the word-excitation guidance, we ask the users to evaluate if the generated gestures have distinct gesticulation at the focus words.
\par
\noindent \textbf{Results.} We plot the results of our user study in~\cref{fig:userstudy}.
For the monadic setting, the participants preferred our motions over those of CaMN and MLD for both questions.
At the same time, we were marginally below the ground-truth preference.
The inference remains similar for the dyadic evaluations as well, although with significantly lower margins.
Finally, the user study demonstrates better semantic alignment with the generated motions with the use of WEG. 
\subsection{Ablative Analysis}
\label{ssec:ablations}
\noindent
\textbf{Latent Representation.} 
\begin{table}
    \centering
    \resizebox{\columnwidth}{!}{
    \begin{tabular}{|c|c|c|} \hline 
         &  Reconstruction Loss $\downarrow$& Smoothness Error~\cite{shimada2020physcap} $\downarrow$\\ \hline 
          MLD \cite{mld}& $10\e{-3}$ &$4.4\e{-3}$\\ \hline 
         Our VAE&  $5\e{-3}$& $3.5\e{-3}$ \\  
         w/o $\mathcal{L}_{lap}$& $3\e{-4}$& $3.7\e{-3}$\\ 
         w/o Time Aware&  $9\e{-3}$& $4\e{-3}$\\ 
         w/o Scale Aware&  $5.5\e{-3}$& $3.7\e{-3}$\\ \hline
    \end{tabular}}
    \caption{\textbf{Ablation study on the VAE design.}
    $\mathcal{L}_{lap}$ ensures the motions retain the velocity of ground truth, even though removing it leads to lower reconstruction loss.
    While training without time-aware representation gives slight increase in reconstruction loss, it cannot support unbounded generation.}
    \label{tab:vae_ablation}
\end{table}
Our chunked, scale-aware latent representation is motivated by various factors, such as perpetual motion synthesis, better temporal alignment with the conditioning modalities, and the scale difference between the hands and the fingers.
We tabulate the influence of the three main design choices in~\cref{tab:vae_ablation}.
We also show that on the VAE reconstruction task alone, our latent representation outperforms MLD's latent representation. 
\begin{figure}
\begin{subfigure}[h]{0.98\linewidth}
\includegraphics[width=\linewidth]{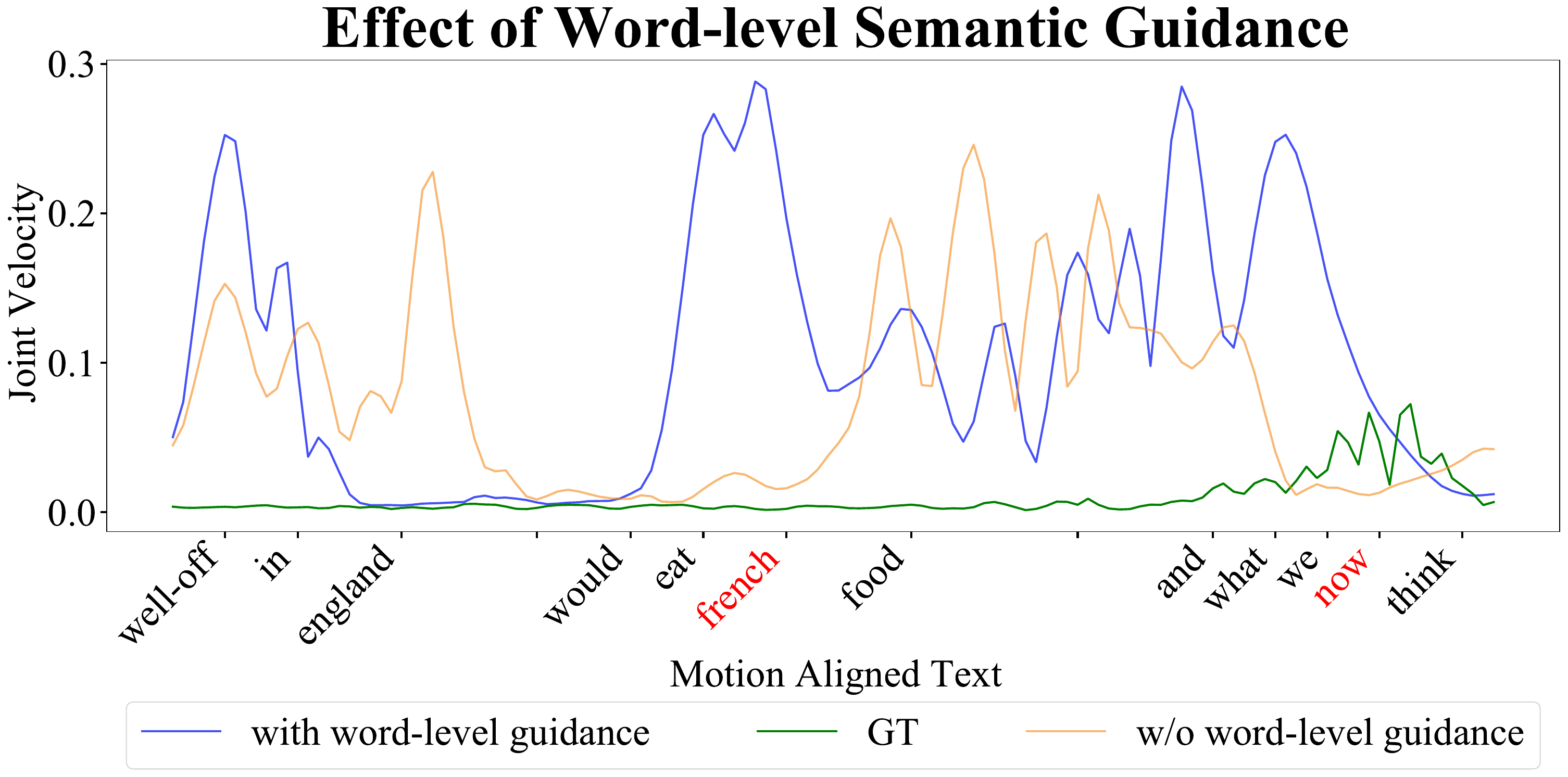}
\caption{Fine-grained alignment} \label{fig:word_excitation}
\end{subfigure}
\hfill
\begin{subfigure}[h]{0.98\columnwidth}
\includegraphics[width=\linewidth]{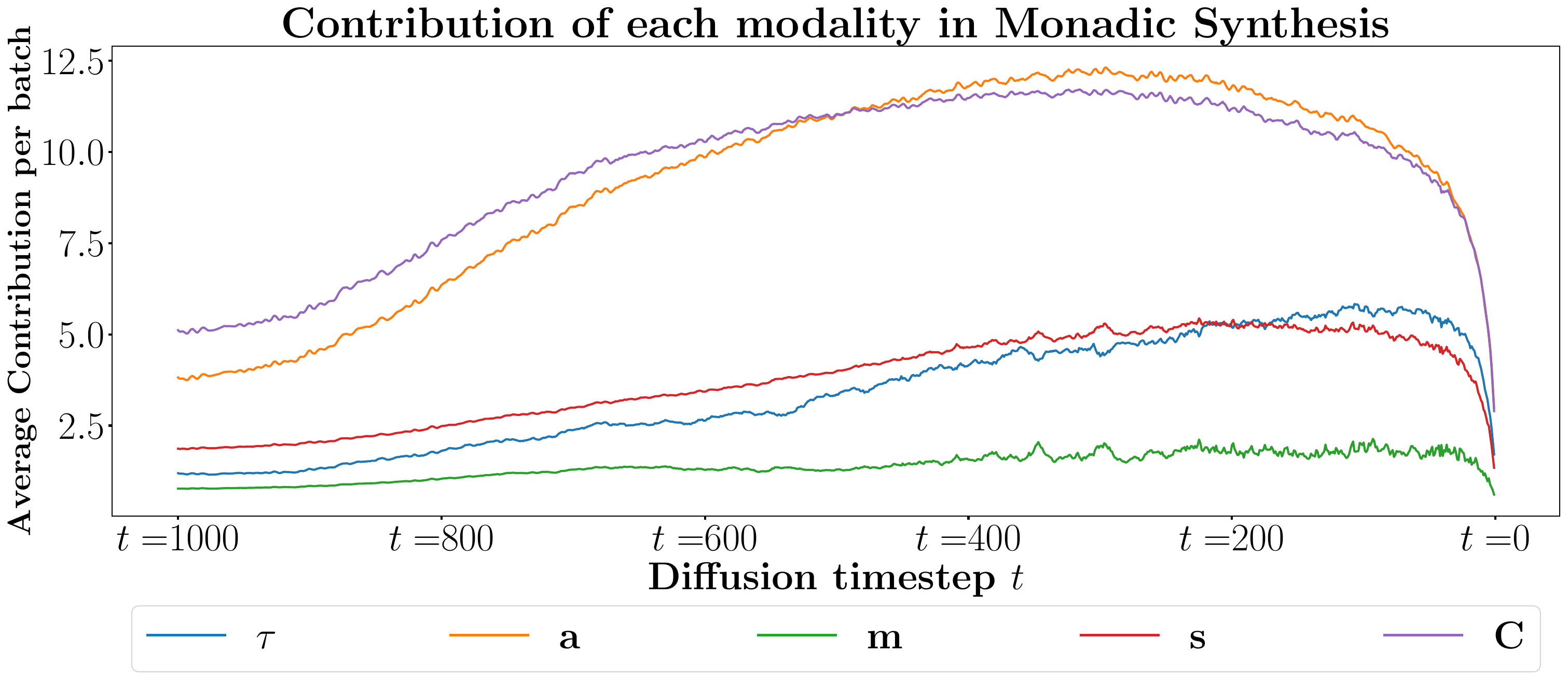}
\caption{The effect of modalities.} \label{fig:modality_effect}
\end{subfigure}%
\caption{(a) Given a text prompt with focus words, ``french'' and ``now'', we observe that WEG significantly increases the joint velocities for the two words compared to the non-guided case. In (b) we show the contributions of each modality as diffusion denoising progresses. Audio tends to dominate the generation process.} \label{fig:overall_plots}
\vspace{-1.0em}
\end{figure}
\par
\noindent
\textbf{Influence of Modalities.} With a variety of conditioning modalities within our framework, it is natural to question which modalities bear a greater effect on the final generation.
We analyze this by plotting the norm of the contributions of each modality in~\cref{eq:modality_guidance} (computed before scaling with $w_c$).
As~\cref{fig:modality_effect} demonstrates, the audio modality bears the largest influence on the gesture generation process. 
Interestingly, we notice an overall trend of increasing contributions until they drop down significantly towards the final stages of denoising, indicating that the diffusion process makes smaller edits in the final stages and takes heavier updates during the middle phases of denoising.
In~\cref{fig:word_excitation}, one can observe a significant bump in the joint velocities (indicating more animated behaviour) at the precise moment of the excitation word.
These observations highlight the overall effectiveness of our two-level guidance objectives. 
We refer the reader to the supplemental for more results.
\par
\noindent
\textbf{On Semantic Consistency:}
Thanks to the proposed Word Excitation Guidance (WEG), our method samples gestures that produce more pronounced attention features for the user-selected words
We demonstrate this by training a gesture type classifier to recognize beat and semantic gestures.
For synthesized gestures without WEG, we observe that the recall for semantic labels is \textbf{0.34}. 
However, this recall increased to \textbf{0.40} when WEG was employed, \textit{indicating that the use of WEG enhances semantic coherence in generated gestures}.
Refer to supplemental for implementation details.
\par
\noindent
\textbf{Attention Maps.}
We visualize the attention maps for analysis (see supplemental) to interpret what spatio-temporal properties are highlighted in the model training.
The first property is a clear separation between the hand and body latents, shown by the striped patterns of the attention maps.
Secondly, WEG boosts the attention weights for the highlighted words. Refer to supplemental for detailed analysis.
%
%
\par
\noindent
\textbf{Perpetual Rollout.}
In addition to allowing for temporal alignment with several modalities, our chunked latent representation also benefits us by allowing perpetual rollout.
To do so, one can simply follow the auto-regressive denoising process followed by the existing motion diffusion methods~\cite{edge, mofusion, mdm} with the difference that instead of inpainting the actual motion, we inpaint the latents. Refer to supplementary material for implementation details.
\section{Conclusion}\label{sec:conclusions}
In this work, we proposed a novel approach towards controllable co-speech gesture synthesis.
With the aim of generating long term, jitter-free gestures, we proposed a time-aware latent representation that can be denoised using a diffusion model.
To control the effects of individual modalities, we proposed a variant of classifier-free guidance.
We also proposed WEG to enhance the gestures for a user-selected set of words in the text, thus facilitating text level fine-grained control. 
Our analysis shows that word-excitation induces more animated behaviour for the selected words. 
Finally, with the introduction of the \dataset dataset 
we hope the field will further propel the research on multi-party gesture synthesis. 
\par
\noindent
\textbf{Acknowledgements.} This work was supported by the ERC Consolidator Grant 4DReply (770784). 
We also thank Andrea Boscolo Camiletto \& Heming Zhu for help with visualizations and Christopher Hyek for designing the game for the dataset.
{
    \small
    \bibliographystyle{ieeenat_fullname}
    \bibliography{references}

\begin{thebibliography}{84}
\providecommand{\natexlab}[1]{#1}
\providecommand{\url}[1]{\texttt{#1}}
\expandafter\ifx\csname urlstyle\endcsname\relax
  \providecommand{\doi}[1]{doi: #1}\else
  \providecommand{\doi}{doi: \begingroup \urlstyle{rm}\Url}\fi

\bibitem[Ahuja et~al.(2020)Ahuja, Lee, Nakano, and Morency]{ahuja2020style}
Chaitanya Ahuja, Dong~Won Lee, Yukiko~I. Nakano, and Louis-Philippe Morency.
\newblock Style transfer for co-speech gesture animation: A multi-speaker conditional-mixture approach.
\newblock 2020.

\bibitem[Alexanderson et~al.(2020)Alexanderson, Henter, Kucherenko, and Beskow]{gen_flow1}
Simon Alexanderson, Gustav~Eje Henter, Taras Kucherenko, and Jonas Beskow.
\newblock Style‐controllable speech‐driven gesture synthesis using normalising flows.
\newblock \emph{Comput. Graph. Forum}, 39\penalty0 (2):\penalty0 487–496, 2020.

\bibitem[Ao et~al.(2022)Ao, Gao, Lou, Chen, and Liu]{rhythm_gest}
Tenglong Ao, Qingzhe Gao, Yuke Lou, Baoquan Chen, and Libin Liu.
\newblock Rhythmic gesticulator: Rhythm-aware co-speech gesture synthesis with hierarchical neural embeddings.
\newblock \emph{ACM TOG}, 41\penalty0 (6):\penalty0 1--19, 2022.

\bibitem[Ao et~al.(2023)Ao, Zhang, and Liu]{gesturediffuclip}
Tenglong Ao, Zeyi Zhang, and Libin Liu.
\newblock Gesturediffuclip: Gesture diffusion model with clip latents.
\newblock \emph{ACM TOG}, 42\penalty0 (4):\penalty0 1--18, 2023.

\bibitem[Bhattacharya et~al.(2021{\natexlab{a}})Bhattacharya, Childs, Rewkowski, and Manocha]{rnn_based1}
Uttaran Bhattacharya, Elizabeth Childs, Nicholas Rewkowski, and Dinesh Manocha.
\newblock Speech2affectivegestures: Synthesizing co-speech gestures with generative adversarial affective expression learning.
\newblock In \emph{ACM MM}, 2021{\natexlab{a}}.

\bibitem[Bhattacharya et~al.(2021{\natexlab{b}})Bhattacharya, Rewkowski, Banerjee, Guhan, Bera, and Manocha]{transformer_based}
Uttaran Bhattacharya, Nicholas Rewkowski, Abhishek Banerjee, Pooja Guhan, Aniket Bera, and Dinesh Manocha.
\newblock Text2gestures: A transformer-based network for generating emotive body gestures for virtual agents.
\newblock In \emph{2021 IEEE Virtual Reality and 3D User Interfaces (VR)}, 2021{\natexlab{b}}.

\bibitem[Busso et~al.(2008)Busso, Bulut, Lee, Kazemzadeh, Mower, Kim, Chang, Lee, and Narayanan]{busso2008iemocap}
Carlos Busso, Murtaza Bulut, Chi-Chun Lee, Abe Kazemzadeh, Emily Mower, Samuel Kim, Jeannette~N Chang, Sungbok Lee, and Shrikanth~S Narayanan.
\newblock Iemocap: Interactive emotional dyadic motion capture database.
\newblock \emph{Language resources and evaluation}, 2008.

\bibitem[Cassell(2000)]{cassell2000embodied}
Justine Cassell.
\newblock Embodied conversational interface agents.
\newblock \emph{Commun. ACM}, 2000.

\bibitem[Cassell et~al.(1994{\natexlab{a}})Cassell, Pelachaud, Badler, Steedman, Achorn, Becket, Douville, Prevost, and Stone]{cassell94animated}
Justine Cassell, Catherine Pelachaud, Norman Badler, Mark Steedman, Brett Achorn, Tripp Becket, Brett Douville, Scott Prevost, and Matthew Stone.
\newblock Animated conversation: Rule-based generation of facial expression, gesture \& spoken intonation for multiple conversational agents.
\newblock In \emph{Proceedings of the 21st Annual Conference on Computer Graphics and Interactive Techniques}, 1994{\natexlab{a}}.

\bibitem[Cassell et~al.(1994{\natexlab{b}})Cassell, Pelachaud, Badler, Steedman, Achorn, Becket, Douville, Prevost, and Stone]{rule1}
Justine Cassell, Catherine Pelachaud, Norman Badler, Mark Steedman, Brett Achorn, Tripp Becket, Brett Douville, Scott Prevost, and Matthew Stone.
\newblock Animated conversation: rule-based generation of facial expression, gesture \& spoken intonation for multiple conversational agents.
\newblock In \emph{SIGGRAPH Conference Proceedings}, 1994{\natexlab{b}}.

\bibitem[Cassell et~al.(2001)Cassell, Vilhj\'{a}lmsson, and Bickmore]{rule2}
Justine Cassell, Hannes~H\"{o}gni Vilhj\'{a}lmsson, and Timothy Bickmore.
\newblock Beat: The behavior expression animation toolkit.
\newblock In \emph{SIGGRAPH Conference Proceedings}, 2001.

\bibitem[Chefer et~al.(2023)Chefer, Alaluf, Vinker, Wolf, and Cohen-Or]{chefer2023attendexcite}
Hila Chefer, Yuval Alaluf, Yael Vinker, Lior Wolf, and Daniel Cohen-Or.
\newblock Attend-and-excite: Attention-based semantic guidance for text-to-image diffusion models.
\newblock \emph{ACM TOG}, 42\penalty0 (4), 2023.

\bibitem[Chen et~al.(2023)Chen, Jiang, Liu, Huang, Fu, Chen, and Yu]{mld}
Xin Chen, Biao Jiang, Wen Liu, Zilong Huang, Bin Fu, Tao Chen, and Gang Yu.
\newblock Executing your commands via motion diffusion in latent space.
\newblock In \emph{CVPR}, 2023.

\bibitem[Dabral et~al.(2023)Dabral, Mughal, Golyanik, and Theobalt]{mofusion}
Rishabh Dabral, Muhammad~Hamza Mughal, Vladislav Golyanik, and Christian Theobalt.
\newblock Mofusion: A framework for denoising-diffusion-based motion synthesis.
\newblock In \emph{CVPR}, 2023.

\bibitem[Dhariwal and Nichol(2021)]{diffusion-beat-gan}
Prafulla Dhariwal and Alexander Nichol.
\newblock Diffusion models beat gans on image synthesis.
\newblock In \emph{NeurIPS}, 2021.

\bibitem[Epstein et~al.(2023)Epstein, Jabri, Poole, Efros, and Holynski]{epstein2023selfguidance}
Dave Epstein, Allan Jabri, Ben Poole, Alexei~A. Efros, and Aleksander Holynski.
\newblock Diffusion self-guidance for controllable image generation, 2023.

\bibitem[Ferstl and McDonnell(2018)]{ferstl18investigating}
Ylva Ferstl and Rachel McDonnell.
\newblock Investigating the use of recurrent motion modelling for speech gesture generation.
\newblock In \emph{Proceedings of the 18th International Conference on Intelligent Virtual Agents}, 2018.

\bibitem[Ferstl et~al.(2020)Ferstl, Neff, and McDonnell]{ferstl2020adversarial}
Ylva Ferstl, Michael Neff, and Rachel McDonnell.
\newblock Adversarial gesture generation with realistic gesture phasing.
\newblock \emph{Computers \& Graphics}, 89:\penalty0 117--130, 2020.

\bibitem[Ghorbani et~al.(2023{\natexlab{a}})Ghorbani, Ferstl, Holden, Troje, and Carbonneau]{zeroeggs}
Saeed Ghorbani, Ylva Ferstl, Daniel Holden, Nikolaus~F. Troje, and Marc‐André Carbonneau.
\newblock Zeroeggs: Zero‐shot example‐based gesture generation from speech.
\newblock \emph{Comput. Graph. Forum}, 42\penalty0 (1):\penalty0 206–216, 2023{\natexlab{a}}.

\bibitem[Ghorbani et~al.(2023{\natexlab{b}})Ghorbani, Ferstl, Holden, Troje, and Carbonneau]{ghorbani2022zeroeggs}
Saeed Ghorbani, Ylva Ferstl, Daniel Holden, Nikolaus~F. Troje, and Marc-André Carbonneau.
\newblock Zeroeggs: Zero-shot example-based gesture generation from speech.
\newblock \emph{Computer Graphics Forum}, 42\penalty0 (1):\penalty0 206--216, 2023{\natexlab{b}}.

\bibitem[Ghosh et~al.(2021)Ghosh, Cheema, Oguz, Theobalt, and Slusallek]{ghosh_2021}
Anindita Ghosh, Noshaba Cheema, Cennet Oguz, Christian Theobalt, and Philipp Slusallek.
\newblock Synthesis of compositional animations from textual descriptions.
\newblock In \emph{International Conference on Computer Vision (ICCV)}, 2021.

\bibitem[Ghosh et~al.(2023)Ghosh, Dabral, Golyanik, Theobalt, and Slusallek]{ghosh2022imos}
Anindita Ghosh, Rishabh Dabral, Vladislav Golyanik, Christian Theobalt, and Philipp Slusallek.
\newblock Imos: Intent-driven full-body motion synthesis for human-object interactions.
\newblock In \emph{Eurographics}, 2023.

\bibitem[Ginosar et~al.(2019)Ginosar, Bar, Kohavi, Chan, Owens, and Malik]{ginosar19learning}
S. Ginosar, A. Bar, G. Kohavi, C. Chan, A. Owens, and J. Malik.
\newblock Learning individual styles of conversational gesture.
\newblock In \emph{Computer Vision and Pattern Recognition (CVPR)}. IEEE, 2019.

\bibitem[Guo et~al.(2022)Guo, Zou, Zuo, Wang, Ji, Li, and Cheng]{humanml3d}
Chuan Guo, Shihao Zou, Xinxin Zuo, Sen Wang, Wei Ji, Xingyu Li, and Li Cheng.
\newblock Generating diverse and natural 3d human motions from text.
\newblock In \emph{CVPR}, 2022.

\bibitem[Habibie et~al.(2021{\natexlab{a}})Habibie, Xu, Mehta, Liu, Seidel, Pons-Moll, Elgharib, and Theobalt]{cnn_based}
Ikhsanul Habibie, Weipeng Xu, Dushyant Mehta, Lingjie Liu, Hans-Peter Seidel, Gerard Pons-Moll, Mohamed Elgharib, and Christian Theobalt.
\newblock Learning speech-driven 3d conversational gestures from video.
\newblock In \emph{Proceedings of the 21st ACM International Conference on Intelligent Virtual Agents}, 2021{\natexlab{a}}.

\bibitem[Habibie et~al.(2021{\natexlab{b}})Habibie, Xu, Mehta, Liu, Seidel, Pons-Moll, Elgharib, and Theobalt]{habibie21learning}
Ikhsanul Habibie, Weipeng Xu, Dushyant Mehta, Lingjie Liu, Hans-Peter Seidel, Gerard Pons-Moll, Mohamed Elgharib, and Christian Theobalt.
\newblock Learning speech-driven 3d conversational gestures from video.
\newblock In \emph{Proceedings of the International Conference on Intelligent Virtual Agents}, 2021{\natexlab{b}}.

\bibitem[Habibie et~al.(2022)Habibie, Elgharib, Sarkar, Abdullah, Nyatsanga, Neff, and Theobalt]{gen_gan}
Ikhsanul Habibie, Mohamed Elgharib, Kripashindu Sarkar, Ahsan Abdullah, Simbarashe Nyatsanga, Michael Neff, and Christian Theobalt.
\newblock A motion matching-based framework for controllable gesture synthesis from speech.
\newblock In \emph{SIGGRAPH ’22 Conference Proceedings}, 2022.

\bibitem[Hendrycks and Gimpel(2016)]{gelu}
Dan Hendrycks and Kevin Gimpel.
\newblock Gaussian error linear units (gelus).
\newblock \emph{arXiv preprint arXiv:1606.08415}, 2016.

\bibitem[Ho and Salimans(2021)]{clf-guidance}
Jonathan Ho and Tim Salimans.
\newblock Classifier-free diffusion guidance.
\newblock In \emph{NeurIPS 2021 Workshop on Deep Generative Models and Downstream Applications}, 2021.

\bibitem[Ho et~al.(2020)Ho, Jain, and Abbeel]{ddpm}
Jonathan Ho, Ajay Jain, and Pieter Abbeel.
\newblock Denoising diffusion probabilistic models.
\newblock In \emph{NeurIPS}, 2020.

\bibitem[Joo et~al.(2017)Joo, Simon, Li, Liu, Tan, Gui, Banerjee, Godisart, Nabbe, Matthews, Kanade, Nobuhara, and Sheikh]{Joo_2017_TPAMI}
Hanbyul Joo, Tomas Simon, Xulong Li, Hao Liu, Lei Tan, Lin Gui, Sean Banerjee, Timothy~Scott Godisart, Bart Nabbe, Iain Matthews, Takeo Kanade, Shohei Nobuhara, and Yaser Sheikh.
\newblock Panoptic studio: A massively multiview system for social interaction capture.
\newblock \emph{IEEE Transactions on Pattern Analysis and Machine Intelligence}, 2017.

\bibitem[Kendon(2004)]{kendon2004gesture}
Adam Kendon.
\newblock \emph{Gesture: Visible action as utterance}.
\newblock Cambridge University Press, 2004.

\bibitem[Kingma and Welling(2014)]{vae}
Diederik~P. Kingma and Max Welling.
\newblock Auto-encoding variational bayes.
\newblock In \emph{ICLR}, 2014.

\bibitem[Kong et~al.(2021)Kong, Ping, Huang, Zhao, and Catanzaro]{diffwave}
Zhifeng Kong, Wei Ping, Jiaji Huang, Kexin Zhao, and Bryan Catanzaro.
\newblock Diffwave: A versatile diffusion model for audio synthesis.
\newblock In \emph{ICLR}, 2021.

\bibitem[Kopp et~al.(2006)Kopp, Krenn, Marsella, Marshall, Pelachaud, Pirker, Th{\'o}risson, and Vilhj{\'a}lmsson]{kopp2006towards}
Stefan Kopp, Brigitte Krenn, Stacy Marsella, Andrew~N Marshall, Catherine Pelachaud, Hannes Pirker, Kristinn~R Th{\'o}risson, and Hannes Vilhj{\'a}lmsson.
\newblock Towards a common framework for multimodal generation: The behavior markup language.
\newblock In \emph{Intelligent Virtual Agents}, 2006.

\bibitem[Kucherenko et~al.(2020)Kucherenko, Jonell, Van~Waveren, Henter, Alexandersson, Leite, and Kjellström]{mlp_based}
Taras Kucherenko, Patrik Jonell, Sanne Van~Waveren, Gustav~Eje Henter, Simon Alexandersson, Iolanda Leite, and Hedvig Kjellström.
\newblock Gesticulator: A framework for semantically-aware speech-driven gesture generation.
\newblock In \emph{Proceedings of the 2020 International Conference on Multimodal Interaction}, 2020.

\bibitem[Kucherenko et~al.(2021{\natexlab{a}})Kucherenko, Nagy, Jonell, Neff, Kjellstr{\"o}m, and Henter]{kucherenko2021speech2properties2gestures}
Taras Kucherenko, Rajmund Nagy, Patrik Jonell, Michael Neff, Hedvig Kjellstr{\"o}m, and Gustav~Eje Henter.
\newblock Speech2{P}roperties2{G}estures: {G}esture-property prediction as a tool for generating representational gestures from speech.
\newblock In \emph{Proceedings of the 21th ACM International Conference on Intelligent Virtual Agents}, 2021{\natexlab{a}}.

\bibitem[Kucherenko et~al.(2021{\natexlab{b}})Kucherenko, Nagy, Jonell, Neff, Kjellstr\"{o}m, and Henter]{semantic1}
Taras Kucherenko, Rajmund Nagy, Patrik Jonell, Michael Neff, Hedvig Kjellstr\"{o}m, and Gustav~Eje Henter.
\newblock Speech2properties2gestures: Gesture-property prediction as a tool for generating representational gestures from speech.
\newblock In \emph{Proceedings of the 21st ACM International Conference on Intelligent Virtual Agents}, 2021{\natexlab{b}}.

\bibitem[Kulkarni et~al.(2023)Kulkarni, Rempe, Genova, Kundu, Johnson, Fouhey, and Guibas]{kulkarni2023nifty}
Nilesh Kulkarni, Davis Rempe, Kyle Genova, Abhijit Kundu, Justin Johnson, David Fouhey, and Leonidas Guibas.
\newblock Nifty: Neural object interaction fields for guided human motion synthesis, 2023.

\bibitem[{Lee} et~al.(2019){Lee}, {Deng}, {Ma}, {Shiratori}, {Srinivasa}, and {Sheikh}]{lee2019talking}
G. {Lee}, Z. {Deng}, S. {Ma}, T. {Shiratori}, S. {Srinivasa}, and Y. {Sheikh}.
\newblock Talking with hands 16.2m: A large-scale dataset of synchronized body-finger motion and audio for conversational motion analysis and synthesis.
\newblock In \emph{2019 IEEE/CVF International Conference on Computer Vision (ICCV)}, 2019.

\bibitem[Levine et~al.(2009)Levine, Theobalt, and Koltun]{stat2}
Sergey Levine, Christian Theobalt, and Vladlen Koltun.
\newblock Real-time prosody-driven synthesis of body language.
\newblock \emph{ACM TOG}, 28\penalty0 (5):\penalty0 1–10, 2009.

\bibitem[Levine et~al.(2010)Levine, Krähenbühl, Thrun, and Koltun]{stat1}
Sergey Levine, Philipp Krähenbühl, Sebastian Thrun, and Vladlen Koltun.
\newblock Gesture controllers.
\newblock \emph{ACM TOG}, 29\penalty0 (4):\penalty0 1–11, 2010.

\bibitem[Li et~al.(2021{\natexlab{a}})Li, Kang, Pei, Zhe, Zhang, He, and Bao]{gen_vae}
Jing Li, Di Kang, Wenjie Pei, Xuefei Zhe, Ying Zhang, Zhenyu He, and Linchao Bao.
\newblock Audio2gestures: Generating diverse gestures from speech audio with conditional variational autoencoders.
\newblock In \emph{ICCV}, 2021{\natexlab{a}}.

\bibitem[Li et~al.(2021{\natexlab{b}})Li, Yang, Ross, and Kanazawa]{aichoreo}
Ruilong Li, Sha Yang, David~A. Ross, and Angjoo Kanazawa.
\newblock Ai choreographer: Music conditioned 3d dance generation with aist++.
\newblock In \emph{ICCV}, 2021{\natexlab{b}}.

\bibitem[Liang et~al.(2022)Liang, Feng, Zhu, Hu, Pan, and Yang]{seeg}
Yuanzhi Liang, Qianyu Feng, Linchao Zhu, Li Hu, Pan Pan, and Yi Yang.
\newblock Seeg: Semantic energized co-speech gesture generation.
\newblock In \emph{CVPR}, 2022.

\bibitem[Liu et~al.(2022{\natexlab{a}})Liu, Iwamoto, Zhu, Li, Zhou, Bozkurt, and Zheng]{disco}
Haiyang Liu, Naoya Iwamoto, Zihao Zhu, Zhengqing Li, You Zhou, Elif Bozkurt, and Bo Zheng.
\newblock Disco: Disentangled implicit content and rhythm learning for diverse co-speech gestures synthesis.
\newblock In \emph{ACM MM}, 2022{\natexlab{a}}.

\bibitem[Liu et~al.(2022{\natexlab{b}})Liu, Zhu, Iwamoto, Peng, Li, Zhou, Bozkurt, and Zheng]{liu2022beat}
Haiyang Liu, Zihao Zhu, Naoya Iwamoto, Yichen Peng, Zhengqing Li, You Zhou, Elif Bozkurt, and Bo Zheng.
\newblock Beat: A large-scale semantic and emotional multi-modal dataset for conversational gestures synthesis.
\newblock \emph{European Conference on Computer Vision}, 2022{\natexlab{b}}.

\bibitem[Liu et~al.(2022{\natexlab{c}})Liu, Wu, Zhou, Du, Wu, Lin, and Liu]{gen_vqvae1}
Xian Liu, Qianyi Wu, Hang Zhou, Yuanqi Du, Wayne Wu, Dahua Lin, and Ziwei Liu.
\newblock Audio-driven co-speech gesture video generation.
\newblock In \emph{NeurIPS}, 2022{\natexlab{c}}.

\bibitem[Liu et~al.(2022{\natexlab{d}})Liu, Wu, Zhou, Xu, Qian, Lin, Zhou, Wu, Dai, and Zhou]{rnn_based2}
Xian Liu, Qianyi Wu, Hang Zhou, Yinghao Xu, Rui Qian, Xinyi Lin, Xiaowei Zhou, Wayne Wu, Bo Dai, and Bolei Zhou.
\newblock Learning hierarchical cross-modal association for co-speech gesture generation.
\newblock In \emph{CVPR}, 2022{\natexlab{d}}.

\bibitem[Loshchilov and Hutter(2019)]{adamw}
Ilya Loshchilov and Frank Hutter.
\newblock Decoupled weight decay regularization.
\newblock In \emph{ICLR}, 2019.

\bibitem[Lugmayr et~al.(2022)Lugmayr, Danelljan, Romero, Yu, Timofte, and Van~Gool]{lugmayr2022diffinpaint}
Andreas Lugmayr, Martin Danelljan, Andres Romero, Fisher Yu, Radu Timofte, and Luc Van~Gool.
\newblock Repaint: Inpainting using denoising diffusion probabilistic models.
\newblock In \emph{CVPR}, 2022.

\bibitem[Marstaller and Burianov{\'a}(2014)]{marstaller2014multisensory}
Lars Marstaller and Hana Burianov{\'a}.
\newblock The multisensory perception of co-speech gestures--a review and meta-analysis of neuroimaging studies.
\newblock \emph{Journal of Neurolinguistics}, 30:\penalty0 69--77, 2014.

\bibitem[McFee et~al.(2015)McFee, Raffel, Liang, Ellis, McVicar, Battenberg, and Nieto]{librosa}
Brian McFee, Colin Raffel, Dawen Liang, Daniel~PW Ellis, Matt McVicar, Eric Battenberg, and Oriol Nieto.
\newblock librosa: Audio and music signal analysis in python.
\newblock In \emph{Proceedings of the 14th python in science conference}, 2015.

\bibitem[McNeill(2008)]{mcneill2008gesture}
David McNeill.
\newblock \emph{Gesture and thought}.
\newblock University of Chicago press, 2008.

\bibitem[Metallinou et~al.(2016)Metallinou, Yang, Lee, Busso, Carnicke, and Narayanan]{metallinou2016usc}
Angeliki Metallinou, Zhaojun Yang, Chi-chun Lee, Carlos Busso, Sharon Carnicke, and Shrikanth Narayanan.
\newblock The usc creativeit database of multimodal dyadic interactions: From speech and full body motion capture to continuous emotional annotations.
\newblock \emph{Language resources and evaluation}, 50:\penalty0 497--521, 2016.

\bibitem[Nyatsanga et~al.(2023)Nyatsanga, Kucherenko, Ahuja, Henter, and Neff]{simba_survey}
S. Nyatsanga, T. Kucherenko, C. Ahuja, G.~E. Henter, and M. Neff.
\newblock A comprehensive review of data-driven co-speech gesture generation.
\newblock \emph{Comput. Graph. Forum}, 42\penalty0 (2):\penalty0 569--596, 2023.

\bibitem[Pang et~al.(2024)Pang, Zhu, Kortylewski, Theobalt, and Habermann]{zhu2023ash}
Haokai Pang, Heming Zhu, Adam Kortylewski, Christian Theobalt, and Marc Habermann.
\newblock Ash: Animatable gaussian splats for efficient and photoreal human rendering.
\newblock In \emph{CVPR}, 2024.

\bibitem[Radford et~al.(2021)Radford, Kim, Hallacy, Ramesh, Goh, Agarwal, Sastry, Askell, Mishkin, Clark, Krueger, and Sutskever]{clip}
Alec Radford, Jong~Wook Kim, Chris Hallacy, Aditya Ramesh, Gabriel Goh, Sandhini Agarwal, Girish Sastry, Amanda Askell, Pamela Mishkin, Jack Clark, Gretchen Krueger, and Ilya Sutskever.
\newblock Learning transferable visual models from natural language supervision.
\newblock In \emph{ICML}, 2021.

\bibitem[Raffel et~al.(2020)Raffel, Shazeer, Roberts, Lee, Narang, Matena, Zhou, Li, and Liu]{raffel2020t5}
Colin Raffel, Noam Shazeer, Adam Roberts, Katherine Lee, Sharan Narang, Michael Matena, Yanqi Zhou, Wei Li, and Peter~J. Liu.
\newblock Exploring the limits of transfer learning with a unified text-to-text transformer.
\newblock \emph{J. Mach. Learn. Res.}, 21\penalty0 (1), 2020.

\bibitem[Ramesh et~al.(2022)Ramesh, Dhariwal, Nichol, Chu, and Chen]{dalle2}
Aditya Ramesh, Prafulla Dhariwal, Alex Nichol, Casey Chu, and Mark Chen.
\newblock Hierarchical text-conditional image generation with clip latents.
\newblock \emph{arXiv}, 2022.

\bibitem[Rombach et~al.(2021{\natexlab{a}})Rombach, Blattmann, Lorenz, Esser, and Ommer]{ldm}
Robin Rombach, Andreas Blattmann, Dominik Lorenz, Patrick Esser, and Björn Ommer.
\newblock High-resolution image synthesis with latent diffusion models.
\newblock In \emph{CVPR}, 2021{\natexlab{a}}.

\bibitem[Rombach et~al.(2021{\natexlab{b}})Rombach, Blattmann, Lorenz, Esser, and Ommer]{stable_diffusion}
Robin Rombach, Andreas Blattmann, Dominik Lorenz, Patrick Esser, and Björn Ommer.
\newblock High-resolution image synthesis with latent diffusion models, 2021{\natexlab{b}}.

\bibitem[Saharia et~al.(2022)Saharia, Chan, Saxena, Li, Whang, Denton, Ghasemipour, Ayan, Mahdavi, Lopes, Salimans, Ho, Fleet, and Norouzi]{imagen}
Chitwan Saharia, William Chan, Saurabh Saxena, Lala Li, Jay Whang, Emily Denton, Seyed Kamyar~Seyed Ghasemipour, Burcu~Karagol Ayan, S.~Sara Mahdavi, Rapha~Gontijo Lopes, Tim Salimans, Jonathan Ho, David~J Fleet, and Mohammad Norouzi.
\newblock Photorealistic text-to-image diffusion models with deep language understanding.
\newblock \emph{arXiv}, 2022.

\bibitem[Shimada et~al.(2020)Shimada, Golyanik, Xu, and Theobalt]{shimada2020physcap}
Soshi Shimada, Vladislav Golyanik, Weipeng Xu, and Christian Theobalt.
\newblock Physcap: Physically plausible monocular 3d motion capture in real time.
\newblock \emph{ACM Transactions on Graphics}, 39, 2020.

\bibitem[Siyao et~al.(2022)Siyao, Yu, Gu, Lin, Wang, Qian, Loy, and Liu]{siyao2022bailando}
Li Siyao, Weijiang Yu, Tianpei Gu, Chunze Lin, Quan Wang, Chen Qian, Chen~Change Loy, and Ziwei Liu.
\newblock Bailando: 3d dance generation via actor-critic gpt with choreographic memory.
\newblock In \emph{CVPR}, 2022.

\bibitem[Sohl-Dickstein et~al.(2015)Sohl-Dickstein, Weiss, Maheswaranathan, and Ganguli]{sohldick}
Jascha Sohl-Dickstein, Eric~A. Weiss, Niru Maheswaranathan, and Surya Ganguli.
\newblock Deep unsupervised learning using nonequilibrium thermodynamics.
\newblock In \emph{ICML}, 2015.

\bibitem[Tanke et~al.(2023)Tanke, Zhang, Zhao, Tang, Cai, Wang, Wu, Gall, and Keskin]{socialdiffusion}
Julian Tanke, Linguang Zhang, Amy Zhao, Chengcheng Tang, Yujun Cai, Lezi Wang, Po-Chen Wu, Juergen Gall, and Cem Keskin.
\newblock Social diffusion: Long-term multiple human motion anticipation.
\newblock In \emph{ICCV}, 2023.

\bibitem[Tevet et~al.(2023)Tevet, Raab, Gordon, Shafir, Bermano, and Cohen-Or]{mdm}
Guy Tevet, Sigal Raab, Brian Gordon, Yonatan Shafir, Amit~H Bermano, and Daniel Cohen-Or.
\newblock Human motion diffusion model.
\newblock In \emph{ICLR}, 2023.

\bibitem[Touvron et~al.(2023)Touvron, Lavril, Izacard, Martinet, Lachaux, Lacroix, Rozière, Goyal, Hambro, Azhar, Rodriguez, Joulin, Grave, and Lample]{touvron2023llama}
Hugo Touvron, Thibaut Lavril, Gautier Izacard, Xavier Martinet, Marie-Anne Lachaux, Timothée Lacroix, Baptiste Rozière, Naman Goyal, Eric Hambro, Faisal Azhar, Aurelien Rodriguez, Armand Joulin, Edouard Grave, and Guillaume Lample.
\newblock Llama: Open and efficient foundation language models, 2023.

\bibitem[Tseng et~al.(2023)Tseng, Castellon, and Liu]{edge}
Jonathan Tseng, Rodrigo Castellon, and Karen Liu.
\newblock Edge: Editable dance generation from music.
\newblock In \emph{CVPR}, pages 448--458, 2023.

\bibitem[Vaswani et~al.(2017)Vaswani, Shazeer, Parmar, Uszkoreit, Jones, Gomez, Kaiser, and Polosukhin]{vaswani2017posenc}
Ashish Vaswani, Noam Shazeer, Niki Parmar, Jakob Uszkoreit, Llion Jones, Aidan~N Gomez, \L~ukasz Kaiser, and Illia Polosukhin.
\newblock Attention is all you need.
\newblock In \emph{NeurIPS}, 2017.

\bibitem[Wagner et~al.(2014)Wagner, Malisz, and Kopp]{wagner_survey}
Petra Wagner, Zofia Malisz, and Stefan Kopp.
\newblock Gesture and speech in interaction: An overview.
\newblock \emph{Speech Communication}, 57:\penalty0 209--232, 2014.

\bibitem[Wolf et~al.(2020)Wolf, Debut, Sanh, Chaumond, Delangue, Moi, Cistac, Rault, Louf, Funtowicz, Davison, Shleifer, von Platen, Ma, Jernite, Plu, Xu, Scao, Gugger, Drame, Lhoest, and Rush]{wolf-etal-2020-transformers}
Thomas Wolf, Lysandre Debut, Victor Sanh, Julien Chaumond, Clement Delangue, Anthony Moi, Pierric Cistac, Tim Rault, Rémi Louf, Morgan Funtowicz, Joe Davison, Sam Shleifer, Patrick von Platen, Clara Ma, Yacine Jernite, Julien Plu, Canwen Xu, Teven~Le Scao, Sylvain Gugger, Mariama Drame, Quentin Lhoest, and Alexander~M. Rush.
\newblock Transformers: State-of-the-art natural language processing.
\newblock In \emph{EMNLP}, 2020.

\bibitem[Yang et~al.(2023)Yang, Wu, Li, Zhang, Hao, Bao, Cheng, and Xiao]{diffstylegest}
Sicheng Yang, Zhiyong Wu, Minglei Li, Zhensong Zhang, Lei Hao, Weihong Bao, Ming Cheng, and Long Xiao.
\newblock Diffusestylegesture: Stylized audio-driven co-speech gesture generation with diffusion models.
\newblock In \emph{IJCAI}, 2023.

\bibitem[Yi et~al.(2023)Yi, Liang, Liu, Cao, Wen, Bolkart, Tao, and Black]{gen_talkshow}
Hongwei Yi, Hualin Liang, Yifei Liu, Qiong Cao, Yandong Wen, Timo Bolkart, Dacheng Tao, and Michael~J Black.
\newblock Generating holistic {3D} human motion from speech.
\newblock In \emph{CVPR}, 2023.

\bibitem[Yoon et~al.(2019{\natexlab{a}})Yoon, Ko, Jang, Lee, Kim, and Lee]{rnn_based4}
Youngwoo Yoon, Woo-Ri Ko, Minsu Jang, Jaeyeon Lee, Jaehong Kim, and Geehyuk Lee.
\newblock Robots learn social skills: End-to-end learning of co-speech gesture generation for humanoid robots.
\newblock In \emph{2019 International Conference on Robotics and Automation (ICRA)}, 2019{\natexlab{a}}.

\bibitem[Yoon et~al.(2019{\natexlab{b}})Yoon, Ko, Jang, Lee, Kim, and Lee]{yoon19robots}
Youngwoo Yoon, Woo-Ri Ko, Minsu Jang, Jaeyeon Lee, Jaehong Kim, and Geehyuk Lee.
\newblock Robots learn social skills: End-to-end learning of co-speech gesture generation for humanoid robots.
\newblock In \emph{Proc. of The International Conference in Robotics and Automation (ICRA)}, 2019{\natexlab{b}}.

\bibitem[Yoon et~al.(2020)Yoon, Cha, Lee, Jang, Lee, Kim, and Lee]{rnn_based3}
Youngwoo Yoon, Bok Cha, Joo-Haeng Lee, Minsu Jang, Jaeyeon Lee, Jaehong Kim, and Geehyuk Lee.
\newblock Speech gesture generation from the trimodal context of text, audio, and speaker identity.
\newblock \emph{ACM TOG}, page 1–16, 2020.

\bibitem[Yuan et~al.(2023)Yuan, Song, Iqbal, Vahdat, and Kautz]{physdiff}
Ye Yuan, Jiaming Song, Umar Iqbal, Arash Vahdat, and Jan Kautz.
\newblock Physdiff: Physics-guided human motion diffusion model.
\newblock In \emph{ICCV}, 2023.

\bibitem[Zhang et~al.(2022)Zhang, Cai, Pan, Hong, Guo, Yang, and Liu]{motiondiffuse}
Mingyuan Zhang, Zhongang Cai, Liang Pan, Fangzhou Hong, Xinying Guo, Lei Yang, and Ziwei Liu.
\newblock Motiondiffuse: Text-driven human motion generation with diffusion model.
\newblock \emph{arXiv preprint arXiv:2208.15001}, 2022.

\bibitem[Zhao et~al.(2023{\natexlab{a}})Zhao, Hu, and Zhang]{diffugesture}
Weiyu Zhao, Liangxiao Hu, and Shengping Zhang.
\newblock Diffugesture: Generating human gesture from two-person dialogue with diffusion models.
\newblock In \emph{International Conference on Multimodal Interaction}, 2023{\natexlab{a}}.

\bibitem[Zhao et~al.(2023{\natexlab{b}})Zhao, Hu, and Zhang]{zhao2023diffugesture}
Weiyu Zhao, Liangxiao Hu, and Shengping Zhang.
\newblock Diffugesture: Generating human gesture from two-person dialogue with diffusion models.
\newblock In \emph{International Cconference on Multimodal Interaction}, pages 179--185. 2023{\natexlab{b}}.

\bibitem[Zhao et~al.(2023{\natexlab{c}})Zhao, Zhou, Li, Tang, Wang, Hou, Min, Zhang, Zhang, Dong, Du, Yang, Chen, Chen, Jiang, Ren, Li, Tang, Liu, Liu, Nie, and Wen]{LLMSurvey}
Wayne~Xin Zhao, Kun Zhou, Junyi Li, Tianyi Tang, Xiaolei Wang, Yupeng Hou, Yingqian Min, Beichen Zhang, Junjie Zhang, Zican Dong, Yifan Du, Chen Yang, Yushuo Chen, Zhipeng Chen, Jinhao Jiang, Ruiyang Ren, Yifan Li, Xinyu Tang, Zikang Liu, Peiyu Liu, Jian-Yun Nie, and Ji-Rong Wen.
\newblock A survey of large language models.
\newblock \emph{arXiv preprint arXiv:2303.18223}, 2023{\natexlab{c}}.

\bibitem[Zhu et~al.(2023)Zhu, Liu, Liu, Qian, Liu, and Yu]{diffgesture}
Lingting Zhu, Xian Liu, Xuanyu Liu, Rui Qian, Ziwei Liu, and Lequan Yu.
\newblock Taming diffusion models for audio-driven co-speech gesture generation.
\newblock In \emph{CVPR}, 2023.

\end{thebibliography}
}
\clearpage
\setcounter{page}{1}
\maketitlesupplementary
This supplementary document provides a glossary of notations used for method explanation in~\cref{supp:glossary} and discusses dataset statistics in~\cref{supp:datastats}. It also provides further details and analyses of word-excitation guidance in~\cref{supp:weg}, user study in~\cref{supp:userstudy} and implementation details in~\cref{supp:implement}. Moreover, we discuss evaluation metrics in~\cref{supp:metrics} and training details for baseline methods in~\cref{supp:baselines}.
\section{Glossary for Notations}
\label{supp:glossary}
In~\cref{tab:supp-vars}, we provide a list of variables used in our the method and implementation details (~\cref{sec:method} and~\cref{supp:implement}) for ease of reference.

\begin{table}[h]
    \centering
    \begin{tabular}{|c|c|} \hline 
         \textbf{Variable}& \textbf{Description}\\ \hline 
         $\mathbf{x}$& Gesture Sequence\\ \hline 
 $\mathbf{x}_b$, $\mathbf{x}_h$&Body and Hand Motions\\ \hline 
         $\mathbf{C}$& Conditioning Set\\ \hline 
         $\mathbf{z}$& Latent representation\\ \hline 
         $\mathbf{z}_b$, $\mathbf{z}_h$& Latent Representation for body and hands\\\hline
 $\xi_b$, $\xi_h$&Encoder for body and hands\\\hline
 $\mathcal{D}_b$, $\mathcal{D}_h$&Decoder for body and hands\\\hline
 ${\mathbf{x}^{\prime}}_b$, ${\mathbf{x}^{\prime}}_h$&Reconstructed motion for body and hands\\\hline
 $\mathbf{\hat{z}}$&Time-aware Latent Representation\\\hline
 $\epsilon_\theta$&Predicted noise\\\hline
 $f_\theta$&Denoiser neural network\\\hline
 $\mathbf{a}$&Audio Signal\\\hline
 $\boldsymbol{\tau}$&Text Embedding\\\hline
 $\boldsymbol{\tau}^{\prime}$&Text Embedding for co-participant\\\hline
 $\mathbf{s}$&Speaker Identity Token\\\hline
 $\mathbf{m}$&Active/Passive Bits for Latent Chunks\\\hline
 $w_c$&Modality guidance scale for condition $\mathbf{c}$\\\hline
 $S$&Number of tokens selected for WEG\\\hline
 $G_{exc}$&Word Excitation Guidance objective\\\hline
 $\mathbf{\tilde{z}^{(t)}}$&Updated latent after WEG\\\hline
    \end{tabular}
    \caption{\textbf{List of variables and their corresponding explanation}}
    \label{tab:supp-vars}
\end{table}
\section{Dataset Statistics \& Discussion}
\label{supp:datastats}
\begin{figure}[t]
    \centering
    \includegraphics[width=\linewidth]{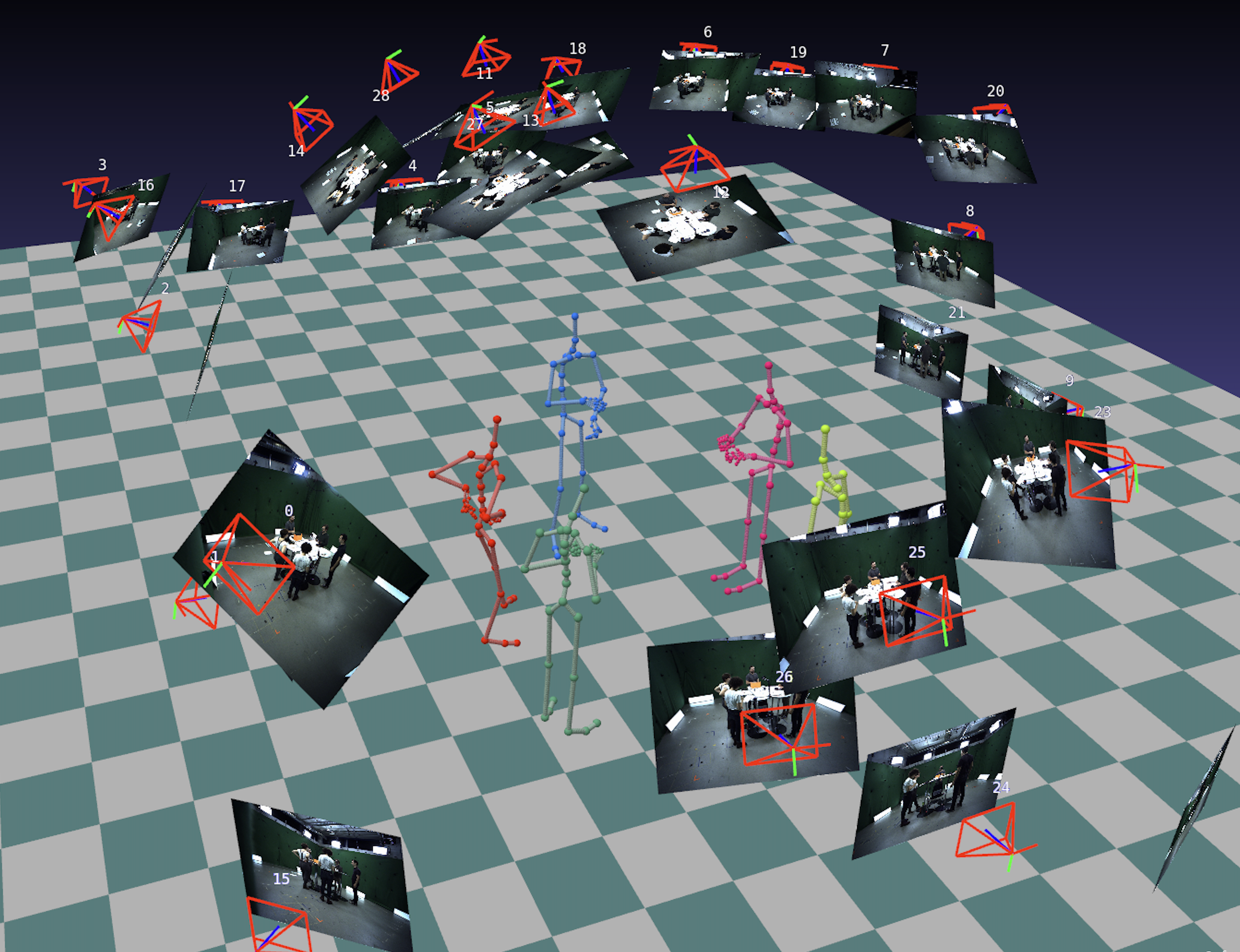}
    \caption{Here, we show an over-arching view of our data-recording setup, where we have five people interacting with each other, while their motion tracking is recorded via a state-of-the-art marker-less motion capture system. Each person also has individual microphones which feed into our audio setup.}
    \label{fig:dataset-aerial}
\end{figure}
The proposed \dataset consists of $6$ hours of mocap data comprising of $5$ persons in the scene.
In total, we have 2.7M poses along with synchronized, per-person audio tracks and text transcripts (see~\cref{fig:dataset-aerial}).
The proposed dataset addresses a different aspect of human gestures, \textit{i.e.} group conversations, which is a sparsely researched setting.
This makes our dataset complementary to the existing monadic gesture datasets like BEAT.
We discuss how BEAT can be used with our framework along with our dataset in~\cref{ssec:baseline-train}.
\par
Finding a capture setting that elicits high density of meaningful, semantic gestures is indeed a challenging task.
These considerations lead us to capture the participants in a role-playing setting as they need to describe an imaginary world to each other, thereby leading to a high density of semantic gestures.
The setting also offers a clear intrinsic reward to the participants (of winning the game).
As we show in the video and~\href{https://vcai.mpi-inf.mpg.de/projects/ConvoFusion/}{website}, the gestures in our dataset are similar to the ones appearing in daily conversation because participants are simply discussing a game plan or their next steps in certain situations using language that is colloquially used in conversations. 
Interestingly, the most relevant gestures to the game setting are pointing gestures (participants usually point to objects on table) which are considered deictic gestures, which happen frequently in normal conversations.
We highlight that the proposed dataset is recorded in a markerless motion capture setup which it keeps the group conversation natural without restrictions of a capture suite or markers, thereby reducing the Observer's Paradox.
Finally, the subjects are familiar with each other (esp. in the DnD setting) which further helps in more natural conversations.
\begin{figure*}
    \centering
    \includegraphics[width=0.8\linewidth]{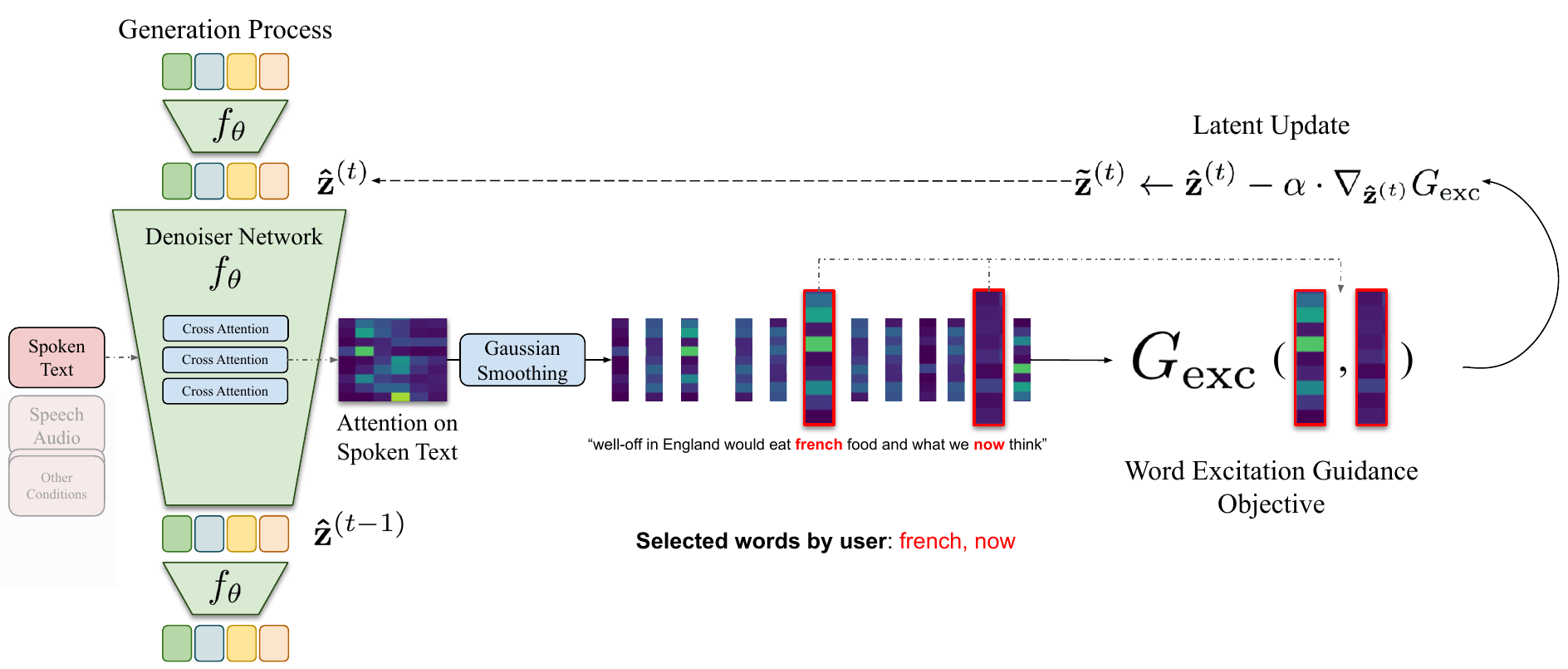}
    \caption{\textbf{Algorithm overview of Word Excitation Guidance.} Here we show the process for the example shown in~\cref{fig:teaser}.}
    \label{fig:word2joints}
\end{figure*}
\section{On Word Excitation Guidance}
\label{supp:weg}
In the following sections, we provide details of algorithm for word-excitation guidance and then perform additional in-depth analysis on its results.
\subsection{Algorithm Details}
\begin{algorithm}
 \caption{Word-Excitation Guidance}
 \begin{algorithmic}[1]
 \renewcommand{\algorithmicrequire}{\textbf{Input:}}
 \renewcommand{\algorithmicensure}{\textbf{Output:}}
 \Require Set of tokens $\{\boldsymbol{\tau}_i\}_{i=1}^{S}$ \\ 
          Trained Diffusion Model $f_\theta$ \\
          Text Prompt $\boldsymbol{\tau} \in \mathbf{C}$, \\
          Diffusion Timesteps $T$ \\
          Step size $\alpha$
 \Ensure  Denoised Latent $\mathbf{\hat{z}}^{(0)}$
 \\ \textit{Initialize} $\mathbf{\hat{z}}^{(T)} \sim \mathcal{N}(0, \mathbf{I})$ 
  \For {$t = T~to~0$} 
      \State $\_, A = f_\theta(\mathbf{\hat{z}}^{(t)}, t, \{\boldsymbol{\tau}\})$ \Comment{get attention for text}
      \State $A \leftarrow \operatorname{Softmax}(A_{\text{start}:\text{end}})$ \Comment{remove start/end of text}
      \State $A \leftarrow \operatorname{Gaussian}(A)$ \Comment{smooth out attentions}
      \State $G_{\mathrm{exc}} = \frac{1}{S} \sum_{i=1}^{S} (1 - \operatorname{max}(A_i))$ \Comment{calculate loss}
i      \State $\mathbf{\Tilde{z}}^{(t)} \leftarrow \mathbf{\hat{z}}^{(t)} - \alpha \cdot \nabla_{\mathbf{\hat{z}}^{(t)}}G_{\mathrm{exc}}$ \Comment{update latent}
      \State \textit{Perform Iterative refinement}~\cite{chefer2023attendexcite}
      \State $\epsilon_{\theta}^{(t)}, \_ = f_{\theta}(\mathbf{\Tilde{z}}^{(t)}, t, \mathbf{C})$ \Comment{estimate noise}
      \State \textit{Perform Modality Guidance}
      \State $\mathbf{\hat{z}}^{(t-1)} \leftarrow \operatorname{SchedulerStep}(\mathbf{\Tilde{z}}^{(t)}, \epsilon_{\theta}^{(t)})$
  \EndFor \\
 \Return  $\mathbf{\hat{z}}^{(0)}$ 
 \end{algorithmic} 
 \label{algo:semantic}
 \end{algorithm}
%
%
%
The process of word-excitation guidance involves modifying the usual denoising loop by updating the latents at each timestep.
Before updating the latents, we normalize the attention maps by removing attention on the start and end tokens because each training batch contains text prompts of different lengths.
Moreover, we observe that our latent diffusion framework assigns high attention to the start token in text (shown in Figure~\ref{fig:attmaps}), therefore, we mitigate this issue by considering the attention on the actual text tokens.
Then we apply Gaussian smoothing over the remaining attention map for stable generation results without any jerks in motion.
This ensures flexibility to focus on a neighbourhood of words instead of one word by avoiding gradient updates at only the chosen tokens ignoring its neighbourhood.
Next, we calculate the average for the loss over all the \textit{focus} tokens to equally transfer gradients for all the focused words.
Note that, this is different from image-based semantic guidance~\cite{chefer2023attendexcite}, where Chefer~\etal apply smoothing on attention for only the chosen words/tokens which ignores the neighbourhood tokens. 
Moreover, their loss aggregation only enables gradient transfer for tokens with the lowest attention instead of all focused tokens by using a $\operatorname{max}$ function instead of $\operatorname{mean}$ like us.
\begin{figure}
    \centering
    \includegraphics[width=\linewidth]{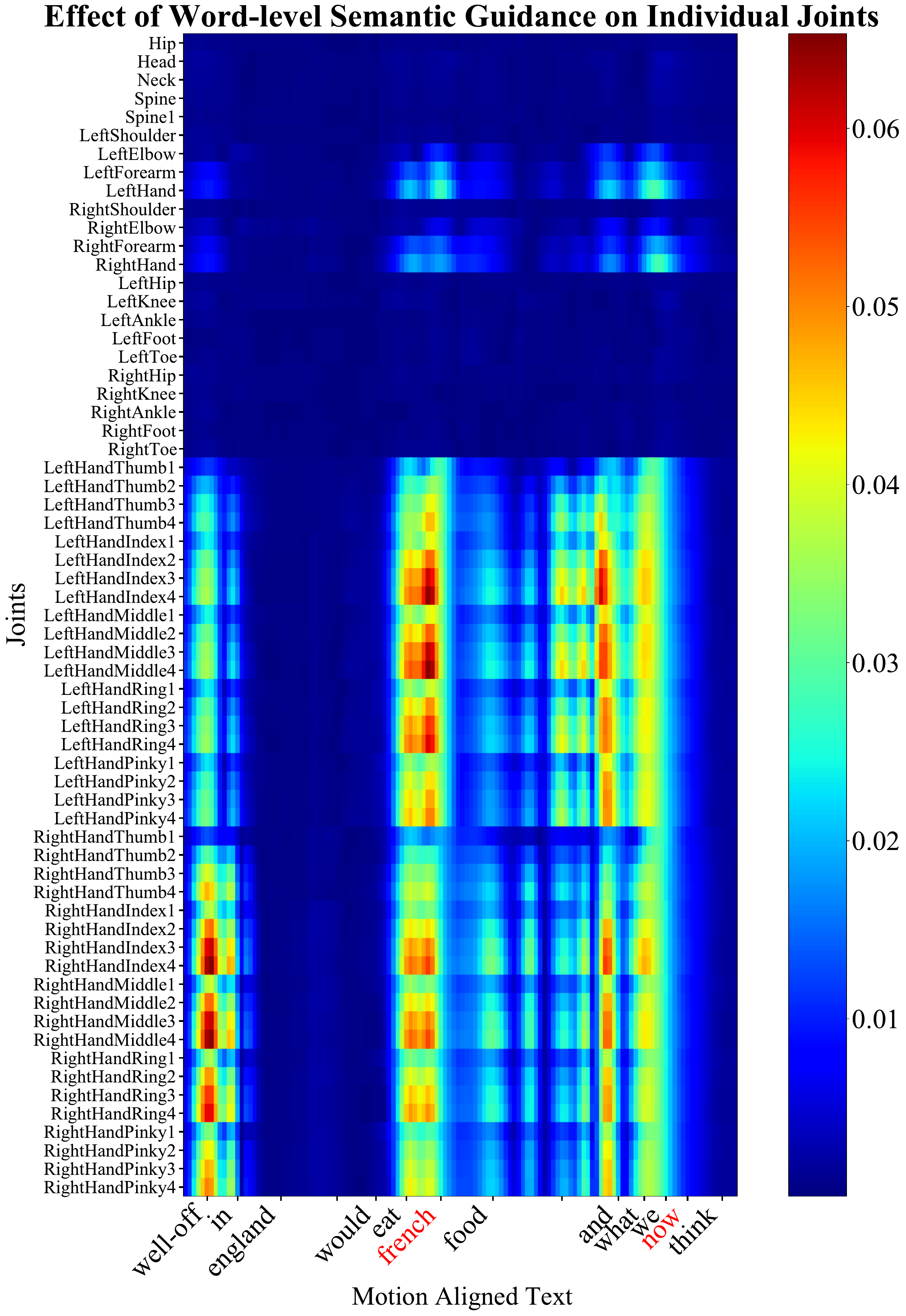}
    \caption{Heatmap showing the distribution of velocity of all joints compared with motion-aligned text (focused words are highlighted in red). We see high joint velocity for hand and arm joints around the words ``french'' and ``now''.}
    \label{fig:word2joints}
\end{figure}
The complete process is presented in Algorithm~\ref{algo:semantic}.
\subsection{Additional Analysis}
\noindent
\textbf{Joints Affected Per Word.}
Recall that word-level excitation guidance steers the gesture generation process through the denoising network to have pronounced gestures at certain words in the text. 
It gives a fine mechanism for semantic control over gesture generation. In~\cref{fig:word2joints}, we present an analysis of how this mechanism affects each joint in the generation. The figure encodes as heatmap the velocity of each joint in response to the text tokens; the assumption being that high velocity implies heavier gesturing. We see that the hand and the arm joints are affected the most at the focused words. Interestingly, minimal attention is focused on the lower body; this is expected as most gestures are predominantly upper-body motions.
\par
\noindent
\textbf{Choice Of Words.} 
Specific types of gestures tend to correlate with certain linguistic structures and parts of speech.
Thus, for analysis, we conduct experiments with attention focused on these elements. 
To extract phrases that may map onto a semantic gesture, we select \textit{random} three-word phrases in the text. 
To experiment with individual words, we can focus on nouns and verbs as they have a higher chance of mapping onto \textit{iconic} gestures. 
Adverbs and adjectives can also be chosen since they can convey spatiotemporal properties of events and entities.
This choice mechanism, which is motivated by the mapping of gestures to linguistic structures, is also flexible enough for the users to choose different linguistic features to focus on. 
We also consider optimal stress word discovery as a future endeavour.
Lastly, the success of word-level semantic guidance is also affected by the amount of stress certain word has in the audio.
We show attention results for phrases and words in~\cref{fig:attmaps} and gesture generation results in supplementary video.
\par
\noindent
\textbf{Interpreting Word-Excitation through Attention Maps.}
Since we perform word-level guidance on text attention maps, we include example results in~\cref{fig:attmaps}.
The attention map $A$ is dependent on text conditioning and diffusion timestep and shows the relation between chunks of latent representation and text tokens.
Therefore, performing word-level guidance at each diffusion timestep yields slightly different attention distribution over words.
We observe that as we move from $t=T$ to $0$, focused word tokens (highlighted in red) start to get high attention, especially after $t=T/2$.
We also see the effect of Gaussian smoothing as the attention on focus words is not sharply focused on only those words.
Rather it is spread over its neighbouring tokens as well.
Lastly, notice the striped pattern of the attention weights. 
This arrangement is a manifestation of the separated body and hand latents that have been stacked alternatively.
It shows that the network learns to perform attention in a separate manner for both types of latents and guidance affects them differently across different layers as well.
\begin{figure*}
\begin{subfigure}[h]{0.98\textwidth}
\includegraphics[width=\linewidth]{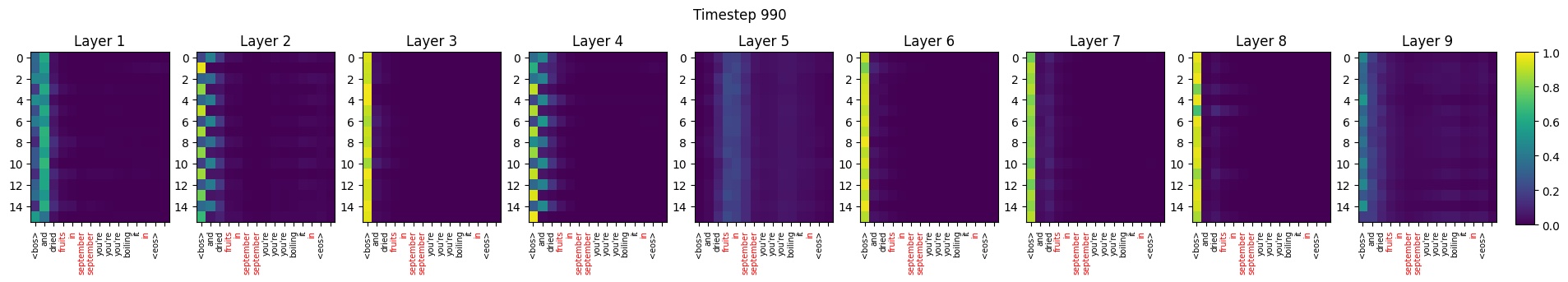}
\caption{Text: ``and dried \textcolor{red}{fruits in september} you're boiling it in''} \label{fig:attmap-b990}
\end{subfigure}
\hfill
\begin{subfigure}[h]{0.98\textwidth}
\includegraphics[width=\linewidth]{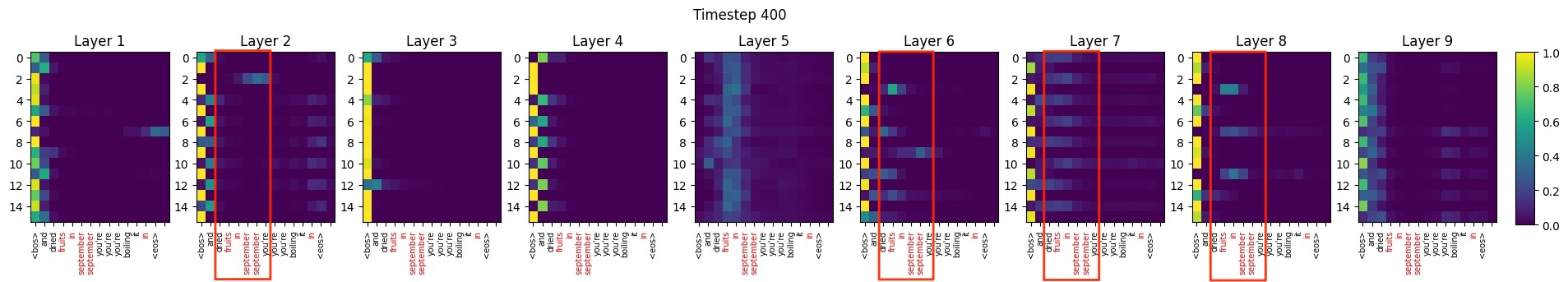}
\caption{Text: ``and dried \textcolor{red}{fruits in september} you're boiling it in''} \label{fig:attmap-b400}
\end{subfigure}%
\hfill
\begin{subfigure}[h]{0.98\textwidth}
\includegraphics[width=\linewidth]{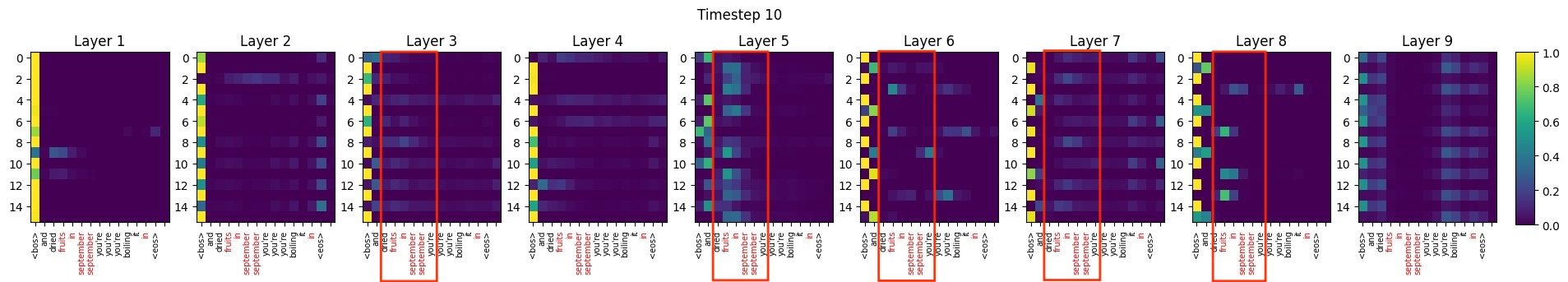}
\caption{Text: ``and dried \textcolor{red}{fruits in september} you're boiling it in''} \label{fig:attmap-b10}
\end{subfigure}
\begin{subfigure}[h]{0.98\textwidth}
\includegraphics[width=\linewidth]{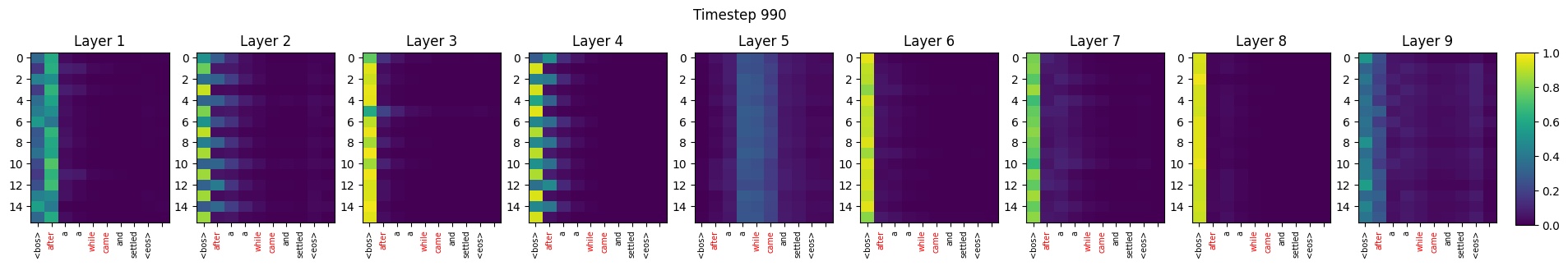}
\caption{Text: ``\textcolor{red}{after} a \textcolor{red}{while came} and settled''} \label{fig:attmap-c990}
\end{subfigure}
\hfill
\begin{subfigure}[h]{0.98\textwidth}
\includegraphics[width=\linewidth]{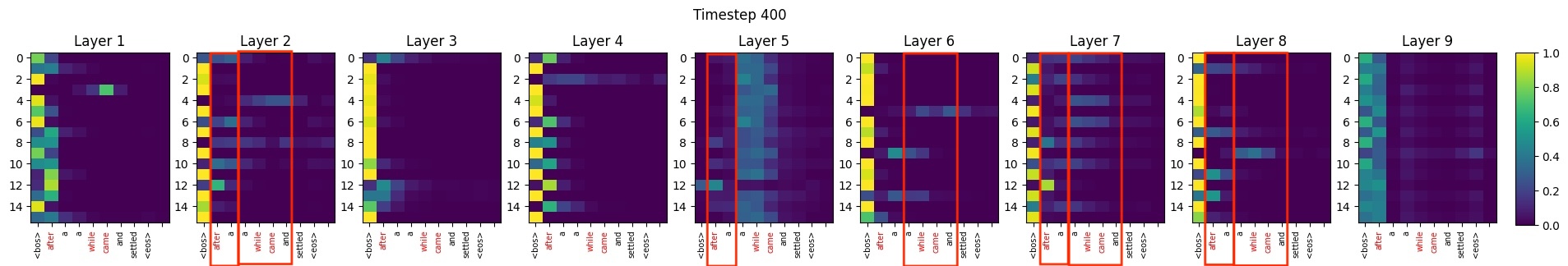}
\caption{Text: ``\textcolor{red}{after} a \textcolor{red}{while came} and settled''} \label{fig:attmap-c400}
\end{subfigure}%
\hfill
\begin{subfigure}[h]{0.98\textwidth}
\includegraphics[width=\linewidth]{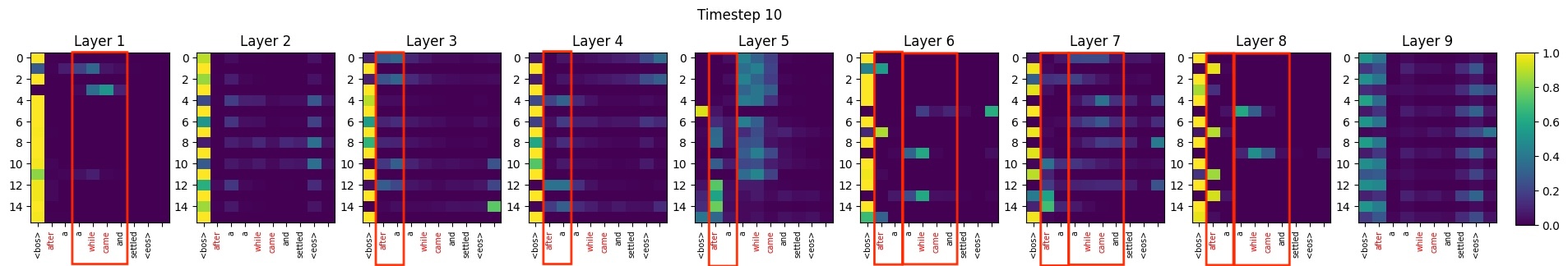}
\caption{Text: ``\textcolor{red}{after} a \textcolor{red}{while came} and settled''} \label{fig:attmap-c10}
\end{subfigure}
\caption{\textbf{Text attention example with focus on a phrase: (a) $t=990$ (b)  $t=400$ (c) $t=10$ and example with multiple individual words: (d) $t=990$ (e)  $t=400$ (f) $t=10$.} Vertical axes show $M \times 2$ latent chunks where even and odd indices stand for body and hand joints respectively. Horizontal axes show word tokens where focused words are highlighted in red. Attention changes are highlighted in red boxes where neighboring tokens are also included to show the effect of Gaussian smoothing. Lastly, ``\textless bos\textgreater'' \& ``\textless eos\textgreater'' tokens represent start and end of the text (refer~\cref{ssec-m:guidance-control})}
\label{fig:attmaps}
\end{figure*}
\par
\noindent
\textbf{Limitations.}
Our method is, after all, a data-driven method.
It depends on the learned conditional gesture distribution of text and other modalities, which can lead to it generating the most common gesture type (\textit{beat gestures}) seen for some words.
Consequently, performing word-level guidance does not always guarantee the specific motion of accurate semantic sub-gesture type (\textit{iconic, deictic, metaphoric etc.}) at the focused word or phrase.
However, as we analyzed, the usage of word-excitation guidance (WEG) mostly results in a semantically meaningful gesture as compared to the base prediction without WEG.
For future works, a more explicit representation of gesture types and their mapping to words can be provided as a conditioning, which might help in predicting semantically accurate gestures.
Secondly, the amount of focus each word/phrase attains in terms of gesture movements is dependent on the fact that speech also contains certain prosodic stress for that word. 
Similarly, if the gestures around focused words are already stressed adequately in motion or those words already have high attention on them, then the change introduced by guidance will only be subtle.
Lastly, the choice of words affects the type of stress in gesture movements predicted and this can be highly subjective. 
\section{User Study}
\label{supp:userstudy}
For evaluation of monadic synthesis, the user were shown a randomly sampled set of $10$ forced-choice questions.
Each question included a side-by-side animation of our method along with one of MLD~\cite{mld}, CaMN~\cite{liu2022beat}, or the ground-truth.
The participants had to answer two question, (a) \textit{``Which of the two gesture motions appears more natural?} and (b) \textit{``Which of the two gesture motions corresponds better with the spoken utterance?''}. 
These questions try to gauge plausibility of the motions and alignment of the generated gestures with the utterance.
For the task of dyadic synthesis, we showed $5$ randomly sampled forced choice questions to each participant, comparing our method with the adapted MLD and the ground-truth.
The participants had to judge the naturalness of the motions similar to previous task and also answer the question: \textit{``In which of the two interactions, the motion of interacting character fits well with both speech of the the main agent as well as their own speech, if any.''}
We report percentage preference for both tasks.
In the third section, we asked the users to evaluate the word-excitation guidance proposed in~\cref{ssec-m:guidance-control}.
Each question included three motions---corresponding to the ground-truth motion, non-guided motion, and word-excitation guided motion --- as well as the words that need to be excited during synthesis.
We compare ground-truth, non-guided gesture, and word-excitation guided gesture.
The users were asked to rate each motion on a Likert scale of $1$-$5$, with $5$ indicating the most semantically aligned gesture.
\section{Implementation Details}
\label{supp:implement}
\textbf{Motion Representation.} 
The motion $\mathbf{x}$ corresponds to the root-relative 3D coordinates for all $J-1$ joints and camera-relative translation of the root joint. 
The hand joints are also made relative to their corresponding wrist joint.
We also pre-process the joint positions following~\cite{humanml3d} by normalizing the motion sequence to start the root trajectory from the origin while facing the positive z-axis.
\paragraph{VAE.}
We implement decoupled scale-aware VAE using two transformer encoders in order to make two halves of the latent representation focus separately on body and hands. 
Each encoder is based on transformer architecture with long skip-connections utilized by Chen \etal~\cite{mld} as they prove this method to be effective in retaining high information density in latent representation.
The output of each encoder is combined into two quantities to represent Gaussian distribution parameters $\boldsymbol{\mu}_\phi$ and $\mathbf{\Sigma}_\phi$ of the combined scale-aware latent space $\mathcal{Z}$, where $\phi$ represent learnable weights of encoders.
We can sample $\mathbf{z}^{2 \times d}$ using reparameterization trick~\cite{vae}.
\par
We train the VAE until convergence with a combination of losses to achieve the desired reconstruction quality.
MSE-based reconstuction loss is applied on the reconstructed motion $\mathbf{\hat{x}}$:
\begin{equation}
    \mathcal{L}_2 = \norm{\hat{\mathbf{x}} - \mathbf{x}}_2
\end{equation}
Moreover, Kullback-Liebler divergence $\mathcal{L}_{KL}$ is used for regularizing the latent space:
\begin{equation}
    \mathcal{L}_{KL} = D_{KL}( \mathcal{N}(\mathbf{z};\boldsymbol{\mu}_\phi, \mathbf{\Sigma}_\phi) || \mathcal{N}(\mathbf{z};0, \mathbf{I}))
\end{equation}
We also apply Bone Length Consistency Loss~\cite{mofusion}, which ensures that bone lengths do not vary across frames in a gesture sequence by minimizing the variance of bone lengths $l_n$.
\begin{equation}
    \mathcal{L}_{bone} = \frac{\sum_{n=1}^N (l_n - \bar{l})^2}{n-1}, 
\end{equation}
Lastly, the VAE loss also contains a Laplacian regularization term $\mathcal{L}_{lap}$ as described earlier, to better reconstruct subtle jerks in gestures and reduce jitter. 
\begin{equation}
    \mathcal{L}_{VAE} = \mathcal{L}_2 + \lambda_{KL} \mathcal{L}_{KL} + \lambda_{lap} \mathcal{L}_{lap} + \lambda_{bone} \mathcal{L}_{bone}
\end{equation}
\par
In order to achieve time-aware latent representation, we encode time-aligned $M$ chunks of motion $\{\mathbf{{x}}_i^{\prime}\}_{i=1}^{M}$ using encoders by passing each $\mathbf{{x}}_i^{\prime}$ from $\xi_b$ and $\xi_h$ to get $\mathbf{\hat{z}}_i$. 
This sequence $\mathbf{\hat{z}} = \{\mathbf{\hat{z}}_i\}_{i=1}^M$ is applied with a positional encoding~\cite{vaswani2017posenc} along $M$ to represent time-alignment.
Along with positional encoding as queries, $\mathbf{\hat{z}}$ passed onto decoder $\mathcal{D}$ as a memory to obtain $\mathbf{\hat{x}}$.
This unique structure of $\mathbf{\hat{z}}$ allows us to perform arbitrary length generation with latent diffusion models, which generally are constrained to due to fixed-length generation. 
\begin{table}[]
    \centering
    \begin{tabular}{|c|c|} \hline  
         \textbf{Hyperparameter}&  \textbf{Value}\\ \hline  
         Latent dimension $d$& 
    128\\ \hline  
 Motion Length $N$&128\\ \hline  
 Number of Joints $J$&63\\ \hline  
 Motion chunks $M$ &8\\ \hline  
 $\lambda_{KL}$&0.05\\ \hline  
 $ \lambda_{lap}$&1\\ \hline
 $\lambda_{bone}$&1\\ \hline  
 Transformer Layers&5\\\hline
 Attention Heads&2\\\hline
 Learning Rate&$1\e{-4}$\\\hline
 Optimizer&AdamW~\cite{adamw}\\\hline
 FPS&25\\\hline \end{tabular}
    \caption{List of values used for training VAE for our method}
    \label{tab:vae_hyperparam}
\end{table}
\paragraph{Perpetual Generation Rollout.}
\begin{figure}
    \centering    
    \includegraphics[width=\columnwidth]{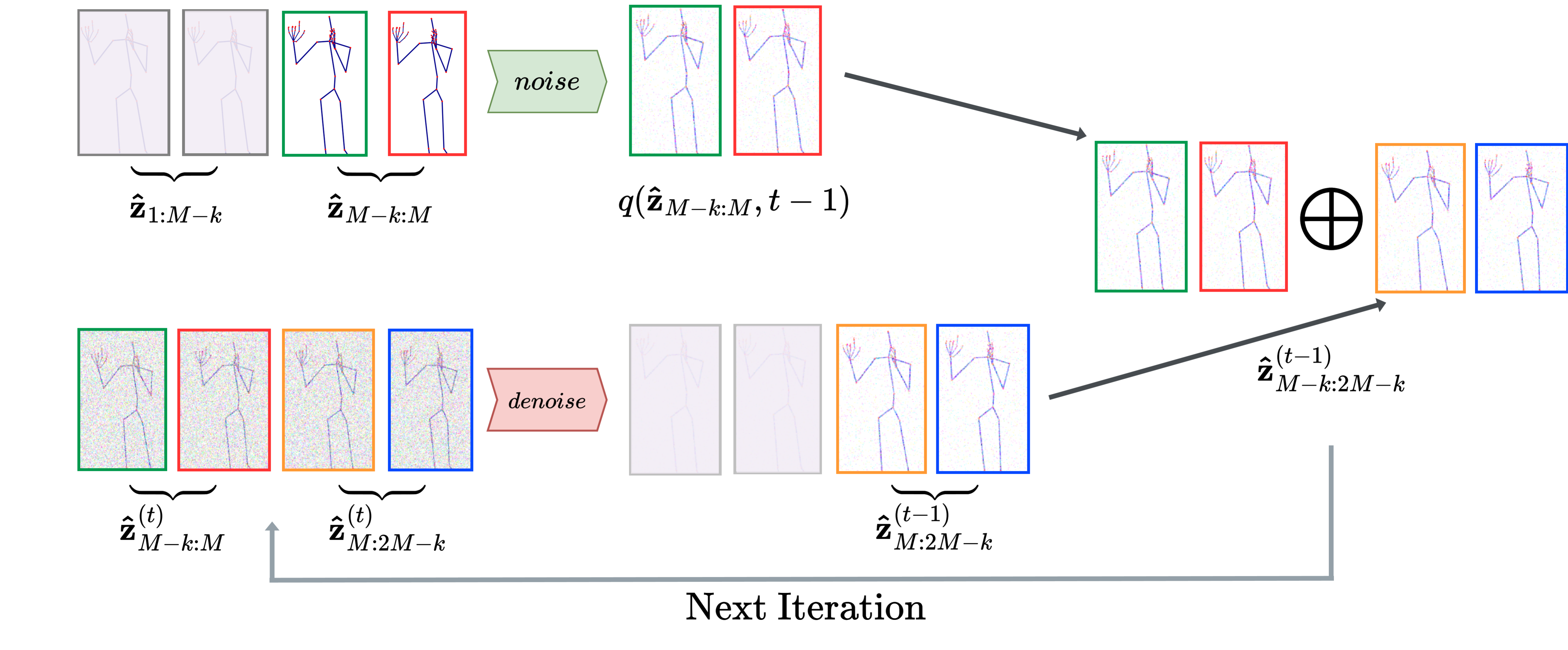}
    \caption{\textbf{Iterative process for the perpetual rollout of arbitrary-length generation.} This is based on the diffusion inpainting technique.}
    \label{fig:inpaint}
\end{figure}
During inference of our diffusion framework, we leverage the time-aware latent sequence $\{\mathbf{\hat{z}}_i\}_{i=1}^M$ to autoregressively generate latent sequences for arbitrarily long sequences.
As compared to earlier approaches \cite{liu2022beat, rnn_based3}, we do not concatenate our model's output which may cause irregular motion at the point of joining. 
We also do not encode variable length sequences in our VAE framework as done by MLD~\cite{mld}.
Instead, we propose an autoregressive generation approach to predict the time-aware latent sequence beyond $M$ number of chunks. 
The key to this approach is an iterative process (shown by~\cref{fig:inpaint}) where a sequence of future latent chunks is predicted based on the last $k$ current latent chunks through a denoising process.
Given the sequence $\mathbf{\hat{z}} \in \mathbb{R}^{M \times 2 \times d}$ represents motion $\mathbf{x}$ of first $N$ frames, we call it $\mathbf{\hat{z}}_{1:M}$ which is known to us.
We utilize the last $k$ latent chunks from this known sub-sequence, i.e. $\mathbf{\hat{z}}_{(M-k):M}$, and generate the next $M-k$ latent chunks through the denoising process to obtain a new overlapping sequence $\mathbf{\hat{z}}_{(M-k):(2M-k)}$. 
Every time we need to generate the next $(M-k):(2M-k)$ sequence, we first inject noise to the previously known $(M-k):M$ sub-sequence until the $t-1$ diffusion timestep. 
Then, this sub-sequence is concatenated with the latent sub-sequence at $M:(2M-k)$ which contains new latents for the non-overlapping part in $(M-k):(2M-k)$ sequence.
This concatenated sequence $\mathbf{\hat{z}}^{(t-1)}_{(M-k):(2M-k)}$ is then passed to the next denoising iteration where the process repeats by noising the known part for the next diffusion timestep.
This technique follows the masked denoising technique used for diffusion image inpainting \cite{lugmayr2022diffinpaint}.
\begin{equation}
	\mathbf{\hat{z}}^{(t-1)}_{M-k:2M-k} = \oplus(q(\mathbf{\hat{z}}_{M-k:M}, t-1), \mathbf{\hat{z}}^{(t-1)}_{M:2M-k})
\end{equation}
Here, the $\oplus$ operator concatenates along latent chunks to total length of $M$ for each sequence.
When applied iteratively to the subsequent new frames, this process enables an autoregressive rollout of fixed-length gesture sequences into infinite-length synthesis. 
We set the value of the hyperparameter $k$ as $k=M/2$ for simplicity.
\paragraph{Details on Denoising Network.}
We design denoising network for the latent diffusion framework to predict $\epsilon_{\theta}(\mathbf{\hat{z}}^{(t)}, t)$. 
We implement the denoising schedule based on DDPM framework with hyperparameters presented in~\cref{tab:diff_hyperparam}.
This framework consists of a Markovian chain of successively adding Gaussian noise $\epsilon$ to $\mathbf{\hat{z}}^{(0)}$ for $T$ timesteps i.e. \textit{forward diffusion} process.
Through this process, $\mathbf{\hat{z}}^{(0)}$, which was sampled from data distribution, becomes $\mathbf{\hat{z}}^{(T)}$, which follows noise distribution $\mathcal{N}(0, \mathbf{I})$ assuming $T$ is sufficiently large.
\begin{equation}
	q \big(\mathbf{\hat{z}}^{(1:T)} | \mathbf{\hat{z}}^{(0)} \big) = \prod_{t=1}^{t=T} q\big(\mathbf{\hat{z}}^{(t)} | \mathbf{\hat{z}}^{(t-1)}\big)
\end{equation}
where $q(\mathbf{\hat{z}}^{(t)} | \mathbf{\hat{z}}^{(t-1)}) = \mathcal{N}(\mathbf{\hat{z}}^{(t)}|\sqrt{1-\beta_t}\mathbf{\hat{z}}^{(t-1)}, \beta_t \mathbf{I})$, describes evolution of latent distribution during the noising process at time step $t$.
Here, $\beta_t$ represents the rate of diffusion. 
The \textit{reverse diffusion} process consists of denoising $\mathbf{\hat{z}}^{(T)} \sim \mathcal{N}(0, \mathbf{I})$ for $T$ timesteps to generate a latent sequence $\mathbf{\hat{z}}^{(0)}$:
\begin{equation}
p_\theta\big(\mathbf{\hat{z}}^{(0:T)}\big) = p\big(\mathbf{\hat{z}}^{(T)}\big) \prod_{t=1}^T p_\theta\big(\mathbf{\hat{z}}^{(t-1)} | \mathbf{\hat{z}}^{(t)}\big),    
\end{equation}
where $p_\theta(\mathbf{\hat{z}}^{(t-1)} |\mathbf{\hat{z}}^{(t)})$ is approximated using a denoiser neural network~$f_{\theta}(\mathbf{\hat{z}}^{(t-1)} | \mathbf{\hat{z}}^{(t)}, t, \mathbf{C})$, which is trained to predict noise
We use transformer decoder network as $f_{\theta}$ which takes $\mathbf{\hat{z}}^{(t-1)}$ as queries along with diffusion timestep $t$ and conditioning set $\mathbf{C}$ as memory input.
We apply positional encoding to queries and individual memory inputs similar to~\cite{vaswani2017posenc}.
To better distinguish between body and hand latents in $\mathbf{\hat{z}}^{(t-1)}$, we add a learned embedding that aims to differentiate between body and hand parts of the latent representation.
We also add a learned embedding to each element of our conditioning set $\mathbf{C}$ separately which helps the network differentiate between different conditioning types.
Each transformer layer starts with Self-Attention and LayerNorm layers, along with a time-layer based on Stylization Block~\cite{motiondiffuse} to incorporate diffusion timestep embedding.
Multi-modal cross attention consists of the same number of heads as the number of elements in the conditioning set $\mathbf{C}$.
The outputs of all heads are aggregated using a linear projection, which is followed by another linear layer with GeLU activation~\cite{gelu}.
\begin{table}[]
    \centering
    \begin{tabular}{|c|c|} \hline  
         \textbf{Hyperparameter}&  \textbf{Value}\\ \hline  
         $d$& 
    128\\ \hline  
 Range of $\beta_t$& $[8.5\e{-4},1.2\e{-2}]$\\ \hline  
 $T$&1000\\ \hline  
 $\beta_t$ Schedule&Scaled Linear~\cite{ldm}\\ \hline  
 Self-Attention Heads&4\\ \hline  
 Decoder Layers&9\\ \hline
 Learning Rate&$7\e{-5}$\\\hline
 Optimizer&AdamW\\\hline \end{tabular}
    \caption{List of hyperparameters for denoising network in our method}
    \label{tab:diff_hyperparam}
\end{table}
\par
\noindent
\textbf{Guidance Parameters.}
We modify classifier-free guidance to add modality-level control for each element in our conditioning set. The random modality dropout rate is set to 10\% and global guidance scale $\lambda_m$ is set to 7.5.
The values of $w_c$ is determined by the task at hand.
For example, if we want to extract only the gesture styles of different speakers regardless of input text and audio, we set all $w_c$ to 0 except $w_{\mathbf{s}} = 1$, which corresponds to speaker identity.
This will generate \textit{unconditional} gestures in the style of a specific speaker (see supplemental video for the example).
For word-excitation guidance, step size $\alpha$, goes from 100 to 70.71 as it varies w.r.t. diffusion timestep.
The kernel size for Gaussian smoothing is 3.
\par
\noindent
\textbf{Semantic Consistency Evaluation Model:}
Our method can generate semantically meaningful gestures (as shown in Suppl. Video), thanks to the proposed Word Excitation Guidance (WEG).
We conduct this ablation the following way.
First, we trained a binary classifier that classifies $1$s motions of the BEAT dataset into either beat gesture, or semantic gesture type (based on the GT labels).
Here semantic class consists of \textit{iconic}, \textit{metaphoric}, and \textit{deictic} classes.
This classifier is then used as an oracle to compute the recall of our generated motions for semantic class predictions.
Specifically, we extract the speech and text for the sentence in which a semantic gesture has been labeled in dataset.
These are then input to \model to generate the corresponding gestures, with and without WEG.
For the case of WEG, we focus on the exact words wherein the semantic gesture occurs in the sentence.
\section{Evaluation Metrics}
\label{supp:metrics}
We report quantitative results on Beat Alignment Score~\cite{aichoreo}, FID, Diversity, L1 Divergence and Semantic Relevance Gesture Recall (SRGR)~\cite{liu2022beat} and here we briefly describe each one of them.
Beat Alignment Score was initially introduced~\cite{aichoreo} to measure the alignment of music beats to dance motion for the task of music-to-dance synthesis. 
This has also been adapted for the task of gesture synthesis, where it measures the correlation between gesture beats and audio beats. 
It is useful in differentiating between static motions which do not align well with the audio from natural-looking gestures which have speech-aligned kinematic beats.
However, it can report false high values if the motion has a large amount of jitter because it would assume beats created by jitter align well with most of the audio beats.
We can see this happening for methods that show high jitter~\cite{rnn_based3, diffgesture} in our experiments. 
They have high Beat Alignment score while their FID is also large. 
\par
We employ the Frechet Inception Distance (FID) metric provided by Yoon \etal~\cite{rnn_based3}, also known as FGD.
We trained our FID network using implementation by Liu~\etal~\cite{liu2022beat}.
It is based on an autoencoder network that is trained for reconstruction task, and is calculated by comparing features of the ground truth data $\mathbf{x}$ and generated data $\mathbf{\hat{x}}$ through:
\par
\footnotesize
\begin{equation}
    \operatorname{FGD}(\mathbf{x}, \mathbf{\hat{x}}) = \norm{\mu_r - \mu_g}^2 + \operatorname{Tr}(\Sigma_r + \Sigma_g - 2 \sqrt{\Sigma_r \Sigma_g} )
\end{equation}
\normalsize
Methods that generate diverse gestures like ours and do not contain pre-pose information unlike CaMN~\cite{liu2022beat} and MultiContext~\cite{rnn_based3}, may suffer on this metric because our gestures will not try to match ground truth motion. 
Diversity computes the average pairwise Euclidean distance of the gesture generations in the test set.
L1 Divergence (also called L1 variance) measures the distance of all frames $N$ in a gesture sequence from their mean $\mu_N$. 
Here $B$ is size of test set.
\begin{equation}
    \operatorname{L1div}(\mathbf{x}) = \frac{1}{B} \sum_{i=1}^N |\mathbf{x}_i - \mu_N|
\end{equation}
This metric specifically identifies if the gestures are static in movement and make less diverse movements along the generation length.
As shown in supplemental video, CaMN~\cite{liu2022beat} and MLD~\cite{mld} suffer from this problem whereas, our method predicts different gestures according to the text and audio conditionings and does not have static motion.
\par
Semantic Relevance Gesture Recall (SRGR) uses semantic score labelled in BEAT~\cite{liu2022beat} as a weight for the Probability of Correct Keypoint (PCK) between the generated gestures and ground truth gestures. It aims to reward being close to ground truth motion at points where a semantically relevant gesture exists while also predicting diverse gestures, as mentioned by Liu \etal~\cite{liu2022beat}.
This metric has a similar issue as FID as it compares PCK between ground truth and prediction and due to many-to-many correspondence between gestures and speech, this might not be suitable. 
Therefore, we conclude that each metric focuses on certain aspects of gesture generation and human-annotated user study results are more conclusive to determine better generation quality and to perform a holistic analysis of gesture quality
\section{Baseline Training Details}
\label{supp:baselines}
In the following, we provide details on how we process all the different modalities for our dataset (\cref{ssec:baseline-datasets}) and provide details for training each method we use for comparison (\cref{ssec:baseline-train}).
\subsection{Dataset}
\label{ssec:baseline-datasets}
\par
\noindent
\textbf{BEAT.} 
We utilize the BEAT dataset~\cite{liu2022beat} in order to augment the training data for our method so that it better generalizes to the task of monadic gesture synthesis.
It consists of 60 hours of English speaking training data, spanning 30 subjects that perform gesture motions.
The dataset is rich in good training examples for monologue setting which can serve as a good baseline dataset to train our method.
Inherently, BEAT's motion representation is different than what we use to train our method.
Therefore, we re-target their skeleton definition to our skeleton definition in order to match it with our \dataset dataset.
Moreover, we convert their representation from BVH-based euler angle representation to joint positions using forward kinematics.
Then we resample their dataset from 120 FPS to 25 FPS in order to match it with our training configuration. 
Lastly, we apply the preprocessing steps mentioned in~\cref{supp:implement} to get the final motions which we separate into 5.12-sec chunks i.e. 128 frames for training our method.
\par
\noindent
\textbf{\dataset.}
To apply our method to the task of dyadic synthesis, we utilize our recorded \dataset dataset (see~\cref{fig:dataset-aerial}) to extract interactions between people in our dataset. 
We record the dataset in BVH format as well, however, we extract joint positions for training our method.
We standardize dataset FPS to 25.
Each of the five people in our dataset has their own separate audio channel which we have post-processed to get clean and denoised audio.
We align audio channels with the recorded tracking and verify it manually as well. 
Finally, we separate out motions for each person and assign them identities which are kept consistent across multiple recording sessions.
Then, we preprocess the dataset and separate it out in chunks of 128 frames.
\par
\noindent
\textbf{Training/Test Splits.}
We split the BEAT dataset by reserving 5 out of 30 English speakers for the testing set, while the remaining go into training and validation splits.
Therefore, all the results and comparisons on monadic synthesis using BEAT dataset are provided on unseen speakers which shows a method's generalizability to unseen audio and text inputs.
For dyadic synthesis, we randomly sample and take out 10 percent for testing and rest for training and validation.
\par
\noindent
\textbf{Representation of Modalities.}
We process audio by sampling it to 16000 Hz and extracting melspectrograms using librosa toolbox~\cite{librosa}.
We use 80 mel-bands and a hop length of 512 for melspectrogram conversion.
We process text through text tokenizer and convert them to embeddings through T5 text encoder~\cite{raffel2020t5} implementation by Hugging Face~\cite{wolf-etal-2020-transformers}.
\par
Lastly, all methods are trained on these dataset splits to ensure fairness. There are some differences between the type of representation used for audio and text in each method, which we elaborate on in the next section.
\subsection{Methods for Comparison}
\label{ssec:baseline-train}

\par
\textbf{ConvoFusion (Ours).}
We train our method on 128-frame sequences by learning a latent space representation of them
Then we use our diffusion framework on top of it.
Interestingly, we can incorporate both monadic and dyadic gesture synthesis tasks into single training.
Thanks to Modality Guidance, we can use different modalities interchangeably by dropping them out of the training batch and setting an unconditional token in their place.
For example, BEAT dataset only contains single-person gesture annotations and does not contain a co-participant, hence making it non-trainable for the task of dyadic synthesis. 
However, we can simply provide an unconditional token for the co-participant's text which automatically turns the contribution of the corresponding modality guidance term to zero, and BEAT dataset can be trained jointly with \dataset dataset.
A similar approach can be taken for semantic annotation labels provided by BEAT dataset, which we do not provide for ours.
\par
\noindent
\textbf{CaMN~\cite{liu2022beat} \& Multi-Context~\cite{rnn_based3}.}
We train both of these methods using the official implementation of CaMN by Liu~\etal on GitHub. 
The only modification that took place was the addition of our dataset pipeline which includes our version of BEAT dataset and \dataset dataset, which makes the motion dimensions from 141 to 189 to match to our setting.
We use the provided $\operatorname{WavEncoder}$ in the implementation to process audio signals instead of melspectrograms.
Lastly, we use motion-aligned text instead of normal text inputs to be consistent with them.
We also use the provided text-encoder to add our textual vocabulary to the text tokenizer for text preprocessing.
\par
\noindent
\textbf{MLD~\cite{mld}.}
This method by Chen~\etal which uses latent diffusion models, was presented for the task of text-to-motion synthesis. We extend this method for the gesture synthesis task by utilizing our training procedure.
To be consistent with their method, we use the text encoder which was used by MLD.
\par
\noindent
\textbf{DiffGesture~\cite{diffgesture}.}
We use their official implementation on GitHub to train this method for the task of monadic gesture synthesis. 
Since DiffuGesture~\cite{diffugesture} does not provide an implementation for dyadic synthesis task and it is highly based on DiffGesture, we follow DiffuGesture's implementation details as close as possible and adapt DiffGesture to the dyadic synthesis task.
As this method was originally trained on TED Dataset~\cite{yoon19robots}, which contains only the upper body, we double the capacity of their transformer network to cater to the increase in dimensionality in our setting to ensure fairness.
The audio and text processing is kept consistent with DiffGesture's implementation. 
Lastly, for the task of dyadic synthesis, we provide co-participant's text as an additional conditioning input to match it with our training pipeline.
%

\end{document}